\documentclass[10pt]{article} 
\usepackage[main, final]{neurips_2026}

\usepackage{natbib}
\setcitestyle{numbers}

\usepackage[utf8]{inputenc} 
\usepackage[T1]{fontenc}    
\usepackage{hyperref}       
\usepackage{url}            
\usepackage{booktabs}       
\usepackage{amsfonts}       
\usepackage{nicefrac}       
\usepackage{microtype}      
\usepackage{xcolor}         

\usepackage{authblk}
\usepackage{amsthm}
\usepackage{amssymb,amsfonts}
\usepackage{graphicx}
\usepackage{textcomp}
\usepackage{authblk}
\usepackage{xcolor}
\newtheorem{definition}{Definition}[section]
\usepackage{comment}
\usepackage[inkscapelatex=false]{svg}
\usepackage{amsmath}
\usepackage{algorithm}
\usepackage{algorithmic}
\usepackage{multirow}
\usepackage{fontawesome5}
\usepackage{wrapfig}

\newtheorem{theorem}{Theorem}
\newtheorem{lemma}{Lemma}
\newtheorem{assumption}{Assumption}
\newtheorem{main_assumption}{Assumption}
\newtheorem{main_theorem}{Theorem}

\newtheorem{corollary}{Corollary}
\newtheorem{proposition}{Proposition}
\newtheorem{main_corollary}{Corollary}

\newcommand{\reals}{\mathbb{R}}
\newcommand{\norm}[1]{\left\|#1\right\|}

\newcommand{\calX}{\mathcal{X}}
\newcommand{\calH}{\mathcal{H}}
\newcommand{\calN}{\mathcal{N}}

\DeclareMathOperator*{\argmax}{arg\,max}

\DeclareMathOperator{\detf}{det}

\newcommand{\bx}{\mathbf{x}}
\newcommand{\by}{\mathbf{y}}
\newcommand{\bk}{\mathbf{k}}
\newcommand{\bK}{\mathbf{K}}
\newcommand{\bC}{\mathbf{C}}
\newcommand{\bI}{\mathbf{I}}

\newcommand{\bepsilon}{\boldsymbol{\varepsilon}}

\title{Bidirectional Information Flow (BIF) - A Sample Efficient Hierarchical Gaussian Process for Bayesian Optimization}

\author[1, 2, 3]{Juan D. Guerra \textsuperscript{\faEnvelope[regular], }}
\author[1, 2]{Thomas Garbay}
\author[3]{\\Numa Dancause}
\author[2, 3]{Guillaume Lajoie \textsuperscript{\faEnvelope[regular], }}
\author[1, 2, 3]{Marco Bonizzato \textsuperscript{\faEnvelope[regular], }}

\affil[]{Polytechnique Montréal $^2$Mila - Québec Artificial Intelligence Institute \newline$^3$Université de Montréal}
\affil[ ]{\textsuperscript{\faEnvelope[regular]}Correspondence: \texttt{juan.guerra@mila.quebec, guillaume.lajoie@mila.quebec, marco.bonizzato@polymtl.ca}}

\begin{document}

\maketitle

\begin{abstract}
Hierarchical Gaussian Process (H-GP) models divide problems into different subtasks, allowing for different components to address each part, making them well-suited for problems with inherent compositional structure. However, existing H-GP frameworks typically employ one-way information sharing — either top-down or bottom-up — which limits sample efficiency and slows convergence. We propose Bidirectional Information Flow (BIF), which establishes continuous two-way communication. BIF retains the modular structure of hierarchical models — the parent conditions its own posterior on child summaries, treating them as structured priors — while introducing top-down feedback to softly decompose environment observations from the parent into sub-responses. This mutual exchange improves sample efficiency, enables robust training, and allows modular reuse of learned subtask models. We prove analytically the regret of a GP with a learned kernel scales linearly with the mismatch to the true kernel, tightening in the hierarchical case to the sum of child-level errors. Ablation shows removing the downward pathway collapses child $R^2$ by up to 58\%. Across synthetic, neurostimulation, and HPO benchmarks, BIF achieves up to 4× higher parent $R^2$ and $\sim$100\% AUC improvement over vanilla GPBO, and outperforms or is the runner-up in 38/40 tests compared to the state-of-the-art baselines while supporting modular child transfer to novel composite tasks.

\end{abstract}

\section{Introduction}

Consider a Bayesian Optimization (BO) problem where the objective decomposes into interacting subtasks: each environment query returns an aggregate signal, but the individual subtask responses are never directly observed. This scenario is widespread - neurostimulation protocols aggregate responses across electrode channels \cite{bonizzato_autonomous_2023}, drug combination effects aggregate over molecular targets \cite{ruberg_application_2023}, and flight control systems couple engine and aerodynamic responses \cite{smith_optimizing_1991}. BO with Gaussian Process (GP) surrogates is well-suited to such settings, offering principled uncertainty quantification, invaluable in high-stakes and low-data regimes such as those mentioned above. However, the key challenge is not merely optimizing the aggregate objective, but doing so efficiently by exploiting the compositional structure.

Hierarchical Gaussian Process (H-GP) models address this by decomposing the objective: \textit{child} GPs model subtasks, and a \textit{parent} GP combines their outputs \citep{laferriere_hierarchical_2020, an_hierarchical_2021}. Nevertheless, existing H-GP frameworks share a critical limitation: information flows in only one direction. In reality, each aggregate observation is a noisy projection of the children's joint responses - it contains information about every subtask simultaneously. By discarding this signal, unidirectional methods \citep{lawrence_hierarchical_2007,  fyshe_hierarchical_2012, laferriere_hierarchical_2020} leave child models frozen after initialization. This means valuable environment information at the parent level — which captures inter-subtask interactions invisible to any individual child — cannot propagate downward and degrades sample efficiency throughout the hierarchy.

We introduce Bidirectional Information Flow (BIF)\footnote{Code and datasets: refer to \url{https://github.com/JuanDGuerra20/Bidirectional-Information-Flow}}, a hierarchical GPBO framework that establishes two-way communication between parent and child models, maximizing the utility of the problem structure. By allowing the parent and children to continually inform each other, BIF reduces the number of optimization steps needed and improves the learning robustness. BIF proposes three novel mechanisms: \textbf{(1) upward prior maps}, where each child's acquisition map is aggregated into a structured prior for the parent, accelerating early-stage optimization; \textbf{(2) a downward response contribution system}, which decomposes each parent observation into approximated sub-responses via a softmax credit-assignment scheme, enabling children to train continuously from global feedback alone; and \textbf{(3) adaptive data rescaling}, which reconciles the two heterogeneous data streams arriving at each child. Together, these mechanisms allow the entire hierarchy to learn in parallel from a single stream of environment queries.

Our contributions are: \textbf{(1)} a hierarchical GP-BO framework unlocking bidirectional information flow that improves sample efficiency, enables online child training from inferred sub-responses, and supports modular transfer of pretrained child GPs to new composite tasks; \textbf{(2)} a regret analysis for GP-UCB under online kernel adaptation, proving that cumulative regret scales linearly with kernel mismatch and, in the hierarchical case, is upper-bounded by the sum of child kernel errors; and \textbf{(3)} Empirical validation on synthetic benchmarks, real-world neurostimulation data, and HPO tasks, demonstrating up to 4$\times$ higher parent $R^2$ and consistent AUC improvements over compared models.

\begin{figure}
    \centering
    \includegraphics[width=\linewidth]{figures/New_BIF_framework_description.pdf}
    \caption{Bidirectional Information Flow framework: Colour coding demonstrates each child's responsibility for a parent dimension. Arrows indicate the direction of information flow: red for prior map creation (child to parent), and blue for sub-response inference via the contribution system (parent to child). Steps (1–7) summarize Alg. \ref{alg:BIF}.}

    \label{fig:bif_framework}
\end{figure}
\section{Related Work}
\label{sec:related_work}

Early hierarchical approaches, including the Hierarchical GP-LVM \citep{lawrence_hierarchical_2007}, Deep GPs \citep{damianou_deep_2013}, and Multi-task GPs \citep{bonilla_multi-task_2007} capture multi-scale patterns and task correlations to improve data efficiency. Similarly, \citet{fyshe_hierarchical_2012} use low dimensional latent processes to classify human magnetoencephalography (MEG) recordings. Explicit hierarchical GP-BO methods directly address structured optimization problems. \citet{laferriere_hierarchical_2020} developed hierarchical GP-BO specifically for neurostimulation protocols, modelling individual neurostimulation channels as child GPs whose outputs form a mean prior for a parent GP. Other work in the field of hierarchical GP-BO was explored by \citet{hensman_hierarchical_2013} in the field of genomics. In both cases, and in H-GP frameworks more broadly, models strictly use a one-way flow of information \citep{lawrence_hierarchical_2007, fox_hierarchical_2007, fyshe_hierarchical_2012, laferriere_hierarchical_2020}. This limitation restricts adaptability and sample efficiency since parent-level insights cannot refine or inform the child models iteratively. BIF addresses this gap directly by introducing a credit assignment mechanism from parent to child and aggregate prior maps from child to parent, significantly enhancing both efficiency and adaptability.

We observe similarities to our hierarchical approach in the field of bilevel optimization, with the closest neighbour being BILBO \citep{chew_bilbo_2025}, which introduces two-way information sharing to reduce the cost of nested function evaluations. The key architectural distinction is that BILBO hard-constrains the upper-level feasible set based on lower-level solutions, whereas BIF treats lower-level outputs as soft priors that the parent GP can adapt or discard given new observations — preserving flexibility when task structure is misspecified or non-stationary.

On the theoretical side, standard GP-UCB regret bounds assume access to the true kernel \citep{srinivas_gaussian_2012}.  In many real-world applications, the kernel is unknown, rendering this assumption invalid. \citet{bogunovic_misspecified_2021} relax this by bounding regret under fixed kernel misspecification. We extend this framework to continuously adaptive kernels, deriving bounds that track kernel mismatch at each step, mimicking the changing dynamics of real-world problems.

\section{Methodology}

\paragraph{Optimization Problem and Model Description.}
\label{sec:optim_problem}
Consider a function $F(x;G,H) = \phi(G(x),H(x))$ (henceforth written $F(x)$) over a space $X$ such that $F$ combines functions $G$ and $H$ in some arbitrary transformation denoted by $\phi$. We would like to model this function using a GP $f(x) \sim GP(\mu(x), k(x,x';\theta))=p(f(x)|x_n,D_n;\theta)$ with $\mu(x)$ being the mean of the GP for point $x$ and $k(x,x';\theta)$ being the kernel function with parameters $\theta$ that will be updated with Bayesian Optimization (BO) ($\theta$ dropped from the notation for brevity, details in Appendix \ref{app:GPBO}). The fit of function $f(x)$ to $F(x)$ could then be optimized by iteratively sampling $X$ and evaluating $F$ while updating our GP model $f$. In hierarchical approaches the functions $G$, $H$ are modeled by GPs $g,h$, and an approximation to their interactions $\phi$ is leveraged to hasten the optimization procedure over $F$. 

We begin with a two layer hierarchical structure, wherein the first layer contains a singular parent GP and the second layer consists of $n$ different child GPs learning independent tasks. In the context of the above problem, the parent GP aims to fit $f(x)$, while child GPs optimize the fit of $g(x)$ and $h(x)$. This hierarchy is explicitly defined in the additive kernel of the parent 
\begin{equation}
    k_p(x,x')=\sum_s^S k_s(x,x')
\end{equation}
where $S$ denotes the set of $n$ child GPs using the Matern kernel \citep{rasmussen_gaussian_2005}. The parent learns how to combine the outputs of the children in order to solve a complex goal made of interactions between the children's tasks. We present this structure in Figure \ref{fig:bif_framework}.

\paragraph{Query Selection using Acquisition Maps.}
\label{sec:queries}
We are interested in maximizing $F(x)$ by iteratively querying it. GPs with BO are often used to do this, together with an acquisition function which is used to pick the next query point.
Acquisition functions $\alpha:X\rightarrow\mathbb{R}$ exploit a surrogate model’s posterior mean $\mu_t(x)$ and variance $\sigma^2_t(x)$ to quantify the utility of each candidate point. Evaluating the acquisition function over the parameter space yields an acquisition map, in which each location is assigned a score $\alpha(x;\mathcal{D}_t)$, where $\mathcal{D}_t=\{(x_i,y_i)\}_{i=1}^t$ represents the set of observations at iteration t, reflecting its expected value for learning or optimization if selected for querying. The next observation is then chosen by the optimization problem $x_{t+1} = \underset{x\in X}{\mathrm{argmax }}\textbf{ }\alpha(x;\mathcal{D}_t)$. In our experiments, BO uses either the Upper Confidence Bound (UCB), Expected Improvement (EI), or Probability of improvement (PI). In the following sections, we will describe the methods using the UCB acquisition function $\alpha(x; \mathcal{D}_t,\theta_t) =\mu_T(x) + \kappa\frac{\sigma_t(x)}{\sqrt{n_t(x)}}$ due to its simplicity and optimality as proven in \citet{auer_finite-time_2002} and \citet{srinivas_gaussian_2012}. An in-depth description of UCB is provided in Appendix \ref{app:UCB}.

\paragraph{Upward Information Flow - Prior Maps.}
\label{sec:upward_flow}
To efficiently optimize the composite task at the parent level, information from the child models is integrated into the parent’s prior. Each child GP independently constructs an acquisition map using an arbitrary function $\alpha$. These child acquisition maps are averaged into a single prior map $A(x)=\frac{1}{|S|}\sum_s^S\alpha_s(x;\mathcal{D}_t,\theta_s)$. We formally incorporate $A(x)$ into the parent GP by defining a parametric mean function $m_p(x; \theta_p)$ utilizing $A(x)$ as a basis function. Specifically, we augment the input space with an intercept and the aggregated map, such that the basis vector is $\boldsymbol{\phi}(x) = [1, A(x)]^\top$. Following the explicit basis function formulation in \citet{rasmussen_gaussian_2005}, the parent GP is specified as:
\begin{equation}
    f_p(x) \sim \mathcal{GP}\big(\boldsymbol{\phi}(x)^\top \boldsymbol{\theta}_p, k_p(x, x')\big), \text{ where }
    \boldsymbol{\theta}_p \sim \mathcal{N}(\mathbf{b}, \mathbf{B}).
\end{equation}
Here, $\boldsymbol{\theta} = [\theta_{\text{bias}}, \theta_{\text{scale}}]^\top$ represents the linear coefficients for the intercept and the aggregated map, respectively. To implement this as a soft structural prior, we specify the prior mean as $\mathbf{b} = [0, 1]^\top$ and fix the prior covariance to $\mathbf{B} = \mathbf{I}$. This configuration explicitly centers the parent’s initial belief on the aggregated child map $A(\mathbf{x})$, biasing the search toward regions deemed informative by the children. Crucially, by employing a fixed, non-zero covariance $\mathbf{B}$, we avoid a rigid dependency; the parent GP benefits from accelerated convergence in the early stages, yet retains sufficient flexibility to adjust the influence of $A(\mathbf{x})$, should the incoming parent data $\mathcal{D}_p$ contradict the child priors.
\paragraph{Downward Information Flow - Response Contribution.}
\label{sec:downward_flow}

To propagate parent-level observations back to the children, we develop a \textit{response contribution system} -- a principled credit-assignment mechanism that decomposes a single noisy global signal $y_p(x)$ into latent sub-responses for each child GP. The decomposition is carried out in two stages, followed by a weighted partition that conserves the global signal.

In the first stage, we bound the magnitude. For each child $s \in S$, we use an arbitrary utility function, set here to UCB, that encodes both the current predictive belief and residual uncertainty:
\begin{equation}
    \alpha_s(x;\theta_s) = \mu_s(x;\theta_s) + \gamma \cdot \frac{\sigma_s(x;\theta_s)}{\sqrt{n_s(x;\theta_s)}}
    \label{eq:utility}
\end{equation}
where $\mu_s$, $\sigma_s$, and $n_s$ are the posterior mean, standard deviation, and query counter of child $s$ respectively, and the hyperparameter $\gamma > 0$ weighs the uncertainty. Raw utility values are not directly comparable across children because each child operates over a different subspace with a different scale. We therefore map each utility onto a unit-less, bounded scale via normalization by the child's own maximum observed utility:
\begin{equation}
    c_s(x) = \frac{\alpha_s(x)}{\max_{x'} \alpha_s(x')}
    \label{eq:magnitude_bound}
\end{equation}
This ensures $c_s(x) \in [0, 1]$ for all $s$ and $x$, preventing dominance by any single child due to scale differences and bounding the inputs to the subsequent softmax, precluding numerical overflow.

In the second stage, the bounded logits are projected onto the probability simplex via a softmax operator:
\begin{equation}
    w_s(x) = \frac{\exp\!\left(c_s(x)\right)}{\sum_{j \in S} \exp\!\left(c_j(x)\right)}
    \label{eq:softmax}
\end{equation}
yielding a valid probability distribution $\{w_s(x)\}_{s \in S}$ with $\sum_{s \in S} w_s(x) = 1$ for all $x$.

\begin{definition}[Response Contribution]
\label{def:response_contribution}
The sub-response assigned to child $s$ at query point $x$ is:
\begin{equation}
    y_s(x) = w_s(x) \cdot y_p(x)
    \label{eq:sub_response}
\end{equation}
where $w_s(x)$ is defined in Equation~\eqref{eq:softmax}. By construction,
\begin{equation}
    \sum_{s \in S} y_s(x) = y_p(x),
    \label{eq:conservation}
\end{equation}
so the sub-responses \emph{exactly conserve} the global parent signal -- the parent's observation noise is fractionally partitioned rather than duplicated, preventing divergence during kernel optimization.
\end{definition}

The inferred data points $\{(x,\, y_s(x))\}$ are added to each child's training set at every parent query, activating a highly efficient learning pipeline and constituting the primary channel by which information flows from the parent to the children. The theoretical properties of this decomposition - including the vanishing bias under balanced children - are formalized in Proposition~\ref{prop:contribution_bias} and proved in Appendix~\ref{app:prop_proof}. Stability properties of the response contribution system are discussed in Appendix \ref{app:stability_properties}. 

Using the contribution system, we create approximate data points for the children based on information received through the parent. The response contribution mechanism acts as an uncertainty-aware credit assignment heuristic, providing a stable learning signal when only aggregated parent-level feedback is observable. In essence, it creates a simulated environment for the children, allowing each parent data point to trickle-down information, even when the children are not directly querying the real environment.

\paragraph{Adaptive Data Rescaling.}
\label{sec:data_norm}
During training, each child receives data from two heterogeneous streams: true environment observations and contribution-inferred sub-responses. To reconcile their differing scales, we apply independent min-max normalization $T,T':Y\rightarrow[0,1]$, where  $T(y_i) = \frac{y_i - \min(Y)}{\max(Y) - \min(Y)}$, yielding normalized datasets $\hat{D}_n = \{x_i, T(y_i)\}_{i=0}^n$ for each stream separately. This ensures each child extracts coherent learning signal from both sources with the true data's normalization $T$ and approximated data's transformation $T'$ nullifying distributional interference.

\paragraph{Quasi-Linear Additive Assumption.}
\label{sec:additive_assumption}

BIF assumes the parent-child interaction $\phi$ takes a quasi-linear additive form, reflected structurally in the additive parent kernel $k_p=\sum_sk_s$ and the averaged prior map $A(x)=\frac{1}{|S|}\sum_s^S\alpha_s(x)$. We provide extensive testing to empirically evaluate the robustness of BIF when this quasi-linear additive assumption is relaxed through scaling nonlinearities with Rosenbrock functions \cite{surjanovic_rosenbrock_2013}, tailor-made functions, and negative interactions (Appendix \ref{app:robustness}).

\begin{wrapfigure}[22]{R}{0.5\textwidth}
\vspace{-25pt}
    \begin{minipage}{\linewidth}
        \begin{algorithm}[H]
        \caption{ Bidirectional Information Flow. }
        \label{alg:BIF}
        \begin{algorithmic}[1]
            \STATE Initialize all child datasets with $r$ datapoints, define query budget Q
            \FOR{q in Q}
                \STATE Generate prior map \(\alpha_s(x; \mathcal{D}_q) \forall s\in S\)  \COMMENT{S is the set of children.}
                \STATE Create average prior \(A = \frac{1}{|S|}\sum_s^S\alpha_s\)
                \STATE Pass $A$ as prior to the mean for Parent $h$
                \STATE Generate Acquisition Map for Parent $\alpha_p$
                \STATE Select $x_q = \max_x(\alpha_p(x))$ to be the next query (number q) for parent h
                \STATE Query environment to get $y_p(x_q)$ as the true environment response
                \STATE Calculate contributions $c_s(x_q)$ 
                \STATE Partition parent response to children $y_s(x_q) = w_s(x)\cdot y_p(x_q)$
                \STATE Add $x_q \text{ to parent and}, T[y(x_q)]$ to respective child GP's training set
                \STATE Increment appropriate query counter for each model in hierarchy
                \STATE Run training for $t$ steps for all models on respective new datasets
            \ENDFOR
        \end{algorithmic}
        \end{algorithm}
    \end{minipage}
\end{wrapfigure}

\paragraph{Training Bidirectional Information Flow.}
\label{sec:BIF_training}

To begin training, we first define a starting dataset of arbitrary size from the true child environments for the child GP models. This is due to the children requiring rudimentary knowledge of the action space to start predicting contributions and creating prior maps. From here, we loop over the training setup involving parent prior map creation, query selection, response contributions propagated to children, and learning outlined in Algorithm \ref{alg:BIF}. Note that an arbitrary maximum query budget Q is used as the stopping criteria for training and can be left open for continual learning, replaced by a runtime threshold, or by a desired performance target.

\section{Theoretical Analysis}
\label{sec:theoretical_analysis}
\paragraph{Regret Bounds for a Learned Kernel with Additive Structure.}

We now analyze the effect of using a learned kernel within the GP-UCB framework. While standard regret bounds assume access to the true kernel, this assumption does not hold when kernels are adapted online. We therefore derive a regret bound that explicitly accounts for kernel mismatch. This result characterizes how deviations between the learned and true kernels impact cumulative regret, and forms the basis for understanding the theoretical behavior of hierarchical and bidirectional GP models.

\begin{main_assumption}[Bounded RKHS Norm]
\label{ass_main:bounded_rkhs}
The true function $f$ has a bounded norm in the Reproducing Kernel Hilbert Space (RKHS) $\calH_{k^*}$ associated with the true kernel $k^*$. We assume there exists a constant $B > 0$ such that $\norm{f}_{\calH_{k^*}} \leq B$.
\end{main_assumption}

\begin{main_assumption}[Uniformly Bounded Kernel Variance]
\label{ass_main:bounded_kernel_variance}
The diagonal of both the true and learned kernels is uniformly bounded. We assume there exists a constant $\kappa > 0$ such that for all $t \in \{1, \dots, T\}$:
\begin{equation}
     \sup_{\bx \in \calX} k^*(\bx, \bx) \leq \kappa^2 \quad \text{and} \quad \sup_{\bx \in \calX} k_{t-1}(\bx, \bx) \leq \kappa^2
\end{equation}
\end{main_assumption}

\begin{main_theorem}[Regret Bound for GP-UCB with a Learned Kernel]
\label{main_thm:regret_bound}
Under Assumptions \ref{ass_main:bounded_rkhs} and \ref{ass_main:bounded_kernel_variance}, let $\delta \in (0, 1)$ and set $\beta_t$ appropriately (see proof). With probability at least $1 - \delta$, the cumulative regret $R_T$ is bounded by:
\begin{equation}
        R_T \le 2 \sqrt{T \beta_T \left(C\gamma_T(k^*) + \sum_{t=1}^T \delta_{\sigma,t}\right)} + \sum_{t=1}^T \left(2\delta_{\mu,t} + 2\sqrt{\beta_t \delta_{\sigma,t}}\right) 
\end{equation}
where $C = \frac{2}{\log(1+\sigma^{-2})}$, $\beta_T = \max_{t\le T} \beta_t$, and the perturbation bounds $\delta_{\mu,t}$ and $\delta_{\sigma,t}$ depend on the kernel mismatch $\Delta_t$ as shown in lemmas \ref{lemma:posterior_mean_perturbation} and \ref{lemma:posterior_variance_perturbation}.
\end{main_theorem}

\begin{main_corollary}[Regret Bound for GP-UCB with a Learned Hierarchical Kernel]
\label{main_cor:hierarchical_regret}
Under Assumptions \ref{ass:bounded_rkhs} and the Additive version of Assumption \ref{ass:bounded_kernel_variance}, let $\delta \in (0, 1)$ and set $\beta_t$ as defined in the proof of Theorem \ref{main_thm:regret_bound}. With probability at least $1-\delta$, the cumulative regret $R_T$ of the GP-UCB algorithm using the structured kernel $k_{t-1} = k_{1, t-1} + k_{2, t-1}$ is bounded by:
\begin{equation}
    R_{T} \le 2\sqrt{T\beta_{T}(C\gamma_{T}(k^{*}) + \sum_{t=1}^{T}\hat{\delta}_{\sigma,t})} + \sum_{t=1}^{T}(2\hat{\delta}_{\mu,t} + 2\sqrt{\beta_{t}\hat{\delta}_{\sigma,t}})
\end{equation}

where $C=\frac{2}{\log(1+\sigma^{-2})}$, $\beta_{T}=\max_{t \le T}\beta_{t}$, $\gamma_T(k^*)$ is the maximum information gain with the true kernel, and the perturbation bounds $\hat{\delta}_{\mu,t}$ and $\hat{\delta}_{\sigma,t}$ depend on the component-wise kernel mismatches $\Delta_{1, t}$ and $\Delta_{2, t}$.
\end{main_corollary}

Theorem~\ref{main_thm:regret_bound} and Corollary~\ref{main_cor:hierarchical_regret} establish how online kernel adaptation affects convergence, yielding two actionable design principles. Theorem~\ref{main_thm:regret_bound} proves that cumulative regret scales linearly with posterior mean error but only as the square root of posterior variance error — a fundamental asymmetry. Since mean error enters at full weight, biases in the predictive mean are disproportionately costly relative to uncertainty misestimation; for learned-kernel GP-UCB, exploitation fidelity therefore governs regret efficiency more than exploration calibration. Corollary~\ref{main_cor:hierarchical_regret} sharpens this for the hierarchical setting: when the kernel is assembled additively from child components, cumulative regret is upper-bounded by the sum of the individual child-level kernel mismatches. This means that improving any single child directly and measurably reduces the regret ceiling of the entire hierarchy — a result that formally justifies BIF's investment in continuously training child models from parent feedback.

We connect these regret bounds to BIF's two information pathways via two propositions (full formulations and proofs in Appendix~\ref{app:prop_proof}). Proposition~\ref{prop:prior_map} proves that the upward prior map converges uniformly to the true objective at a rate controlled by child prediction error and residual uncertainty: as children improve, the prior map tightens around the true objective and directly accelerates parent-level convergence. Proposition~\ref{prop:contribution_bias} proves that the downward softmax contribution introduces a bounded, characterizable bias that vanishes exactly in equal contribution cases as children converge. Together with Proposition \ref{prop:prior_map}, this provides a convergence argument for both pathways linking to the regret bound of Corollary \ref{main_cor:hierarchical_regret} via reduced child kernel mismatch. This chain of results elevates BIF from a heuristically motivated architecture to a framework with proven guarantees — a property of particular importance in safety-critical domains such as neurostimulation, where reliability is non-negotiable.

\section{Experiments}

\paragraph{Datasets.}
\label{sec:dataset}

We evaluate BIF on five synthetic and three real-world benchmarks designed to test hierarchical transfer and scalability. All synthetic tasks include 10\% additive Gaussian Noise. \textbf{(i) Michalewicz 2D \cite{surjanovic_michalewicz_2013}:}  a synthetic landscape with steep valleys and a narrow global optimum. \textbf{(ii) Holder Table 2D \cite{surjanovic_holder_2013}:} a synthetic 2D function with four global minima and many local minima aimed at distracting the optimizer. \textbf{(iii) 3D Synthetic:} A higher-dimensional task with three children $g(x)=-(x-2)^5$, $h(y)=\sin^3(y)$, and $k(z)=\frac{\log(z+1)+1}{z+1}$. The parent function is $f(x,y,z) = (g(x)+h(y)+k(z))^2 + 2$. This benchmark assesses BIF's robustness as the hierarchy scales. \textbf{(iv) Levy 3D \cite{surjanovic_levy_2013}:} A standard 3D optimization benchmark used to show BIF's ability to perform in higher dimensional problems. \textbf{(v) Hartmann 6D \cite{surjanovic_hartmann_2013}:} aimed to show the effects of higher dimensional scale on BIF as performance drops are expected with more dimensions. \textbf{(vi) Neurostimulation:} A real-world engineering task optimizing electrode channels implanted in the motor cortex of a monkey for maximal muscle activation \citep{laferriere_samlafhierarchical-gaussian-process_2019}. This dataset evaluates BIF's applicability in high-stakes problems, where success could dictate the implementation of BIF to \textit{in vivo} neurostimulation protocols. \textbf{(vii) HyperParameter Optimization (HPO):} These real-world tasks aim to optimize a Lasso model's \cite{tibshirani_regression_1996} learning rate and L1 ratio on sklearn's diabetes dataset \cite{pedregosa_scikit-learn_2011} or a GAN's generator (learning rate and weight decay through one child GP) and discriminator (learning rate and dropout rate through another child GP) on Fashion-MNIST \cite{xiao_fashion-mnist_2017}.
Further tests pertaining to BIF's robustness, mechanistic insights including ablation studies, and significance testing can be found in Appendices \ref{app:robustness}, \ref{app:mechanistic_insights}, and \ref{app:significance} respectively.

\paragraph{Implementation Details.}
\label{sec:BIF_setup}

BIF hyperparameters ($\kappa, \gamma, r, t$) and GP settings are tuned via independent studies in Appendix \ref{app:hp_selection}; for neurostimulation, optimal values are $\{8, 4, 1, 15\}$. We employ two online normalization strategies for the true environment responses, chosen as a hyperparameter according to the dataset, using only available trajectory data: z-score scaling for synthetic tasks or scaling the observations by the maximum response observed so far. We benchmark BIF against vanilla GPBO \citep{rasmussen_gaussian_2005}, Deep GPs \cite{damianou_deep_2013}, Additive GPs (ADD-GP) \cite{duvenaud_additive_2011} and the hierarchical model by \citet{laferriere_hierarchical_2020}. Child GPs employ Matern kernels ($\nu=0.5$), while the parent uses an additive kernel $k_p=\sum_s^S k_s$. All models are updated via the ADAM optimizer ($\alpha=0.01$) \citep{kingma_adam_2017}. Results report the mean and standard error across 10 randomly initialized repetitions.

\begin{figure}
    \centering
    \includegraphics[width=\linewidth]{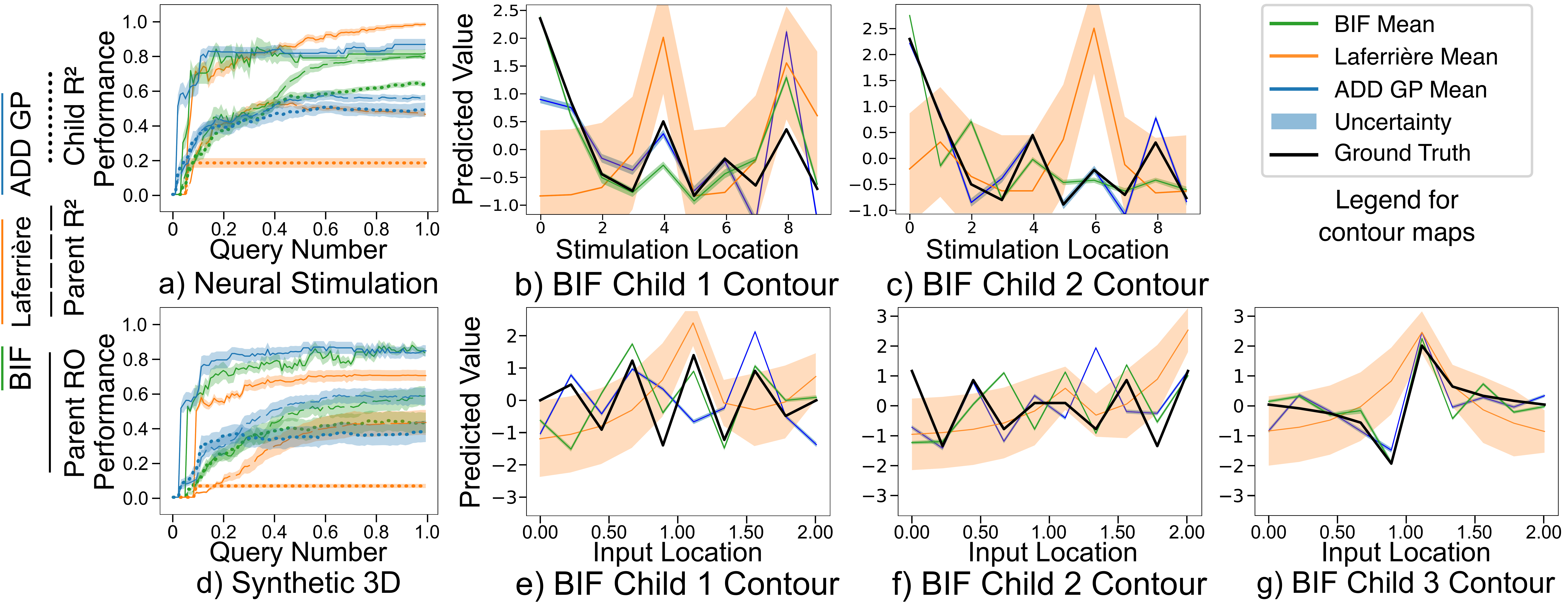}
    \caption{Baseline GPBO models vs. BIF UCB on 2D (a–c) and 3D (d–g) synthetic tasks. (a, d) shows all model performance for the 2D and 3D tasks with shaded regions depicting standard error; (b-c, e-g) show BIF child environment reconstructions (green: BIF; orange: \citet{laferriere_hierarchical_2020}, black: ground truth). See Section~\ref{sec:dataset} for details. Models selected based on presence of direct sub-task modeling for comparison. Child RO metric is omitted for ease of reading. All results presented in Tables \ref{table:performance_synthetic} \& \ref{table:performance_real}. Parent Visualizations for Neural Stimulation available in Appendix \ref{app:parent_heatmap}.}
    \label{fig:main_text_plots}
\end{figure}
\paragraph{Child Process Modularity.}
\label{sec:modularity}
Our model features modular child components that can be reused across tasks. To validate our claim of modularity, we present an experimental protocol involving the pretraining of child GPs on one combination task, to later be switched into another combination task. For instance, take two pretraining optimization tasks, $f_p$ with children $f_a,f_b$ and task $g_p$ with children $g_c,g_d$. We propose an experiment to optimize task $h_p$ with child targets $f_a,g_d$ by inserting the pretrained children from tasks $f_p,g_p$ to bootstrap the performance of task $h_p$. As a baseline, a regular BIF model is trained from scratch to compare the boost in performance seen by injecting the pretrained children to the training protocol. A visual description and further details can be found in Figure \ref{fig:modularity_sketch} and Appendix \ref{app:modularity}.

\paragraph{Performance Metrics.}
\label{sec:performance_metrics}
We report three complementary metrics. The Relative Optimum (RO) measures parent optimization quality: \(
RO = \frac{y(\hat{x}) + y(\tilde{x})}{y(x^*)+ y(\tilde{x})} \text{ such that } \hat{x}=\arg\max \alpha(x;\theta)\) is the predicted optimum, $\tilde{x}$ is the ground truth minimum location, $x^*$ is the ground truth maximum location, and $y$ is the true function. This metric then measures the percent of maximum response attained, similar to regret. The coefficient of determination, $R^2$ measures surrogate fidelity, in other words, how closely the learned GP $f(x)$ captures the true function $F(x)$. Both the RO and $R^2$ are reported separately for parent and as the average across all children. Finally, the Area Under the Curve (AUC), sums all of the above metrics over the training run, rewarding early convergence and sustained performance. Together, these metrics distinguish models that find good optima quickly and model the landscape accurately. 

\begin{table*}[t]
\caption{Summary of synthetic performance after 100 queries of training. Vanilla GPBO and Deep GP do not contain direct subtask mapping, rendering them unable to do the child RO and $R^2$ tasks.}
\label{table:performance_synthetic} 
\centering
\resizebox{\textwidth}{!}{%
\begin{tabular}{ccccccc}
\hline
\textbf{Dataset}                                                              & \textbf{Model} & \textbf{Parent RO $\uparrow$} & \textbf{Parent R² $\uparrow$} & \textbf{Child RO $\uparrow$} & \textbf{Child R² $\uparrow$} & \textbf{AUC $\uparrow$} \\ \hline
\multirow{7}{*}{Michalewicz}                                                  & Vanilla GP     & 97.98 ± 1.52        & 55.29 ± 1.67                     & -                            & -                               & 10,138.74               \\
                                                                              & Deep GP        & 28.46 ± 7.64                  & 25.19 ± 6.38                     & -                            & -                               & 3,356.45                \\
                                                                              & ADD GP         & 97.27 ± 0.71                  & 75.17 ± 1.80                     & 93.25 ± 2.96                 & 75.74 ± 2.34                    & 18,716.24               \\
                                                                              & Laferrière     & 97.85 ± 0.71                  & 70.81 ± 1.93                     & 74.79 ± 14.21                & 18.29 ± 3.71                    & 14,992.09               \\
                                                                              & BIF UCB        & 98.87 ± 0.60                  & \textbf{86.10 ± 0.88}            & \textbf{94.25 ± 2.66}        & 61.88 ± 3.25                    & 18,587.90               \\
                                                                              & BIF EI         & \textbf{98.94 ± 0.52}                  & 84.69 ± 0.99                     & 94.03 ± 6.25                 & 69.63 ± 2.42                    & \textbf{19,226.67}      \\
                                                                              & BIF PI         & 92.11± 4.81                   & 81.56 ± 4.01                     & 92.28 ± 6.16                 & \textbf{76.08 ± 3.14}           & 18,132.09               \\ \hline

\multirow{7}{*}{Holder 2D}                                                    & Vanilla GP     & 96.49 ± 6.68                  & 35.10 ± 12.48                    & -                            & -                               & 9,803.89                \\
                                                                              & Deep GP        & 72.43 ± 29.76                 & 18.65 ± 11.42                    & -                            & -                               & 7,794.28                \\
                                                                              & ADD GP         & 93.60 ± 8.43                  & 53.97 ± 5.85                     & \textbf{96.46 ± 4.71}        & 64.32 ± 6.68                    & 24,221.08               \\
                                                                              & Laferrière     & \textbf{99.34 ± 0.76}         & 64.02 ± 4.66                     & 37.18 ± 22.00                & 5.72 ± 5.42                     & 17,898.00               \\
                                                                              & BIF UCB        & 97.66 ± 4.41                  & \textbf{64.73 ± 6.93}            & 94.29 ± 5.87                 & \textbf{69.50 ± 8.36}           & \textbf{25,448.41}      \\
                                                                              & BIF EI         & 95.27 ± 6.63                  & 47.82 ± 12.76                    & 90.48 ± 15.18                & 50.99 ± 17.19                   & 23,772.08               \\
                                                                              & BIF PI         & 91.42 ± 20.95                 & 47.91 ± 14.74                    & 92.58 ± 15.57                & 44.53 ± 14.70                   & 21,117.74               \\ \hline
\multirow{7}{*}{Synthetic 3D}                                                 & Vanilla GP     & 80.79 ± 0.28                  & 13.39 ± 0.13                     & -                            & -                               & 7,677.74                \\
                                                                              & Deep GP        & 82.21 ± 5.85                  & \textbf{62.94 ± 7.75}            & -                            & -                               & 10,320.54               \\
                                                                              & ADD GP         & 84.82 ± 3.27                  & 58.57 ± 4.91                     & 86.79 ± 11.42                & 38.21 ± 6.23                    & \textbf{15,727.70}      \\
                                                                              & Laferrière     & 70.79 ± 0.32                  & 43.30 ± 0.73                     & 75.62 ± 16.24                & 6.52 ± 0.12                     & 9,473.47                \\
                                                                              & BIF UCB        & 88.24 ± 3.23                  & 61.39 ± 2.55                     & 82.81 ± 12.48                & 42.86 ± 5.87                    & 14,986.14               \\
                                                                              & BIF EI         & \textbf{92.19 ± 2.65}         & 56.75 ± 5.94                     & \textbf{88.52 ± 5.55}        & 38.37 ± 6.12                    & 13,787.51               \\
                                                                              & BIF PI         & 76.45 ± 3.53                  & 60.33 ± 6.46                     & 87.29 ± 10.95                & \textbf{53.01 ± 5.67}           & 13,701.51               \\ \hline
\multirow{7}{*}{Levy 3D}                                                      & Vanilla GP     & 97.10 ± 2.37                  & 2.30 ± 2.76                      & -                            & -                               & 9,573.02                \\
                                                                              & Deep GP        & 90.43 ± 12.62                 & 14.03 ± 11.37                    & -                            & -                               & 9,013.18                \\
                                                                              & ADD GP         & 98.09 ± 1.38                  & \textbf{78.74 ± 12.25}           & 96.36 ± 4.16                 & \textbf{72.72 ± 11.43}          & \textbf{32,259.02}      \\
                                                                              & Laferrière     & 97.66 ± 1.92                  & 49.83 ± 26.88                    & 91.20 ± 14.27                & 5.89 ± 3.27                     & 23,277.83               \\
                                                                              & BIF UCB        & 97.99 ± 1.59                  & 54.38 ± 20.34                    & 95.69 ± 5.54                 & 57.20 ± 12.47                   & 28,768.29               \\
                                                                              & BIF EI         & \textbf{98.28 ± 1.30}         & 31.61 ± 17.71                    & \textbf{96.50 ± 3.69}        & 45.94 ± 12.92                   & 26,528.72               \\
                                                                              & BIF PI         & 97.90 ± 1.92                  & 34.51 ± 14.90                    & 94.96 ± 4.53                 & 49.34 ± 15.45                   & 27,118.25               \\ \hline
\multirow{7}{*}{Hartmann 6D}                                                  & Vanilla GP     & \textbf{86.79 ± 14.86}        & \textbf{23.32 ± 7.19}            & -                            & -                               & 5,618.41                \\
                                                                              & Deep GP        & 7.49 ± 9.03                   & 1.14 ± 2.03                      & -                            & -                               & 665.74                  \\
                                                                              & ADD GP         & 70.73 ± 29.41                 & 8.79 ± 3.50                      & \textbf{89.61 ± 14.73}       & \textbf{61.67 ± 13.83}          & \textbf{21,375.72}      \\
                                                                              & Laferrière     & 19.15 ± 18.08                 & 2.77 ± 2.80                      & 74.40 ± 16.12                & 47.91 ± 14.86                   & 13,893.00               \\
                                                                              & BIF UCB        & 76.01 ± 23.23                 & 6.98 ± 2.54                      & 74.24 ± 26.03                & 59.70 ± 23.26                   & 17,757.59               \\
                                                                              & BIF EI         & 58.36 ± 28.38                 & 4.01 ± 3.19                      & 74.17 ± 25.22                & 48.81 ± 18.64                   & 17,543.12               \\
                                                                              & BIF PI         & 68.31 ± 29.34                 & 4.71 ± 2.79                      & 85.81 ± 19.30                & 59.58 ± 20.72                   & 20,706.06               \\ 
    
\end{tabular}
}
\end{table*}

\begin{table*}[t]
\caption{Summary of real-world performance after 100 queries of training. Vanilla GPBO and Deep GP do not contain direct subtask mapping, rendering them unable to do the child RO and $R^2$ tasks.}
\label{table:performance_real} 
\centering
\resizebox{\textwidth}{!}{%
\begin{tabular}{ccccccc}
\hline
\textbf{Dataset}                                                              & \textbf{Model} & \textbf{Parent RO $\uparrow$} & \textbf{Parent R² $\uparrow$} & \textbf{Child RO $\uparrow$} & \textbf{Child R² $\uparrow$} & \textbf{AUC $\uparrow$} \\ \hline
\multirow{7}{*}{\begin{tabular}[c]{@{}c@{}}Neural\\ Stimulation\end{tabular}} & Vanilla GP     & \textbf{98.46 ± 0.07}         & 70.13 ± 0.11                     & -                            & -                               & 14,119.99      \\
                                                                              & Deep GP        & 78.18 ± 3.33                  & 69.58 ± 2.61                     & -                            & -                               & 11,728.75               \\
                                                                              & ADD GP         & 86.92 ± 3.22                  & 56.03 ± 1.66                     & 86.61 ± 17.27                & 49.31 ± 4.05                    & 17,087                  \\
                                                                              & Laferrière     & 96.37 ± 0.11                  & 45.45 ± 0.15                     & 55.34 ± 26.14                & 17.84 ± 0.27                    & 13,799.99               \\
                                                                              & BIF UCB        & 81.83 ± 0.25                  & 79.70 ± 0.12                     & \textbf{100.00 ± 0.00}       & \textbf{63.98 ± 0.15}           & \textbf{18,299.35}      \\
                                                                              & BIF EI         & 93.20 ± 2.71                  & 60.59 ± 2.20                     & 88.88 ± 15.66                & 37.22 ± 3.92                    & 16,566.67               \\
                                                                              & BIF PI         & 77.36 ± 0.00                  & \textbf{81.43 ± 0.99}            & 99.22 ± 4.29                 & 60.41 ± 2.63                    & 17,660.21               \\ \hline
\multirow{7}{*}{HPO - GAN}                                                    & Vanilla GP     & 96.04 ± 0.45                  & 9.67 ± 1.00                      & -                            & -                               & 2,035.21                \\
                                                                              & Deep GP        & 94.31 ± 0.85                  & 1.53 ± 0.35                      & -                            & -                               & 1,889.41                \\
                                                                              & ADD GP         & 94.71 ± 0.77                  & 9.24 ± 1.34                      & 73.18 ± 19.47                & 26.59 ± 2.81                    & 2,430.96                \\
                                                                              & Laferrière     & 94.72 ± 0.55                  & 6.76 ± 1.08                      & 75.73 ± 16.10                & 23.08 ± 2.42                    & 2,244.58                \\
                                                                              & BIF UCB        & \textbf{96.50 ± 0.45}         & \textbf{12.19 ± 0.86}            & 70.75 ± 17.62                & 26.72 ± 1.74                    & 2,596.53                \\
                                                                              & BIF EI         & 94.81 ± 0.43                  & 9.47 ± 0.82                      & \textbf{76.45 ± 17.05}       & 29.26 ± 2.24                    & 2,596.89                \\
                                                                              & BIF PI         & 94.36 ± 0.44                  & 8.82 ± 0.82                      & 76.15 ± 14.10                & \textbf{30.04 ± 1.92}           & \textbf{2,609.34}       \\ \hline
\multirow{7}{*}{HPO - Lasso}                                                  & Vanilla GP     & 99.82 ± 0.11                  & 43.72 ± 1.40                     & -                            & -                               & 14,139.57               \\
                                                                              & Deep GP        & 99.86 ± 0.04                  & \textbf{89.27 ± 5.89}            & -                            & -                               & 14,655.16               \\
                                                                              & ADD GP         & 99.26 ± 0.17                  & 57.35 ± 1.50                     & 63.06 ± 19.66                & 51.62 ± 4.17                    & 19,572.69               \\
                                                                              & Laferrière     & 99.81 ± 0.09                  & 48.44 ± 3.35                     & 10.82 ± 14.48                & 16.99 ± 3.79                    & 15,302.20               \\
                                                                              & BIF UCB        & \textbf{99.98 ± 0.02}         & 75.35 ± 2.84                     & \textbf{78.49 ± 21.70}       & 50.96 ± 5.79                    & 17,906.14               \\
                                                                              & BIF EI         & 99.44 ± 0.30                  & 67.82 ± 1.94                     & 33.02 ± 29.97                & 54.04 ± 6.91                    & 19,104.27               \\
                                                                              & BIF PI         & 99.83 ± 0.07                  & 79.96 ± 0.94                     & 30.97 ± 28.24                & \textbf{71.63 ± 3.40}           & \textbf{20,249.31}     
\end{tabular}
}
\end{table*}

\section{Results}
\label{sec:results}

\paragraph{Synthetic Tasks (Table \ref{table:performance_synthetic}).}
On Michalewicz, BIF is the undisputed best model. With the EI acquisition function, it reaches the best parent RO scores of 98.94 while simultaneously having the lowest standard error, showing its performance and reliability. On surrogate accuracy, BIF UCB attains the highest parent $R^2$ of any model (86.10 vs. ADD GP: 75.17), attributable to upward prior maps supplying structured early guidance that accelerates kernel learning. Child fidelity reveals the sharpest separation: BIF PI achieves the highest child $R^2$ of any model with 76.08. Comparing this to Laferrière's 18.29, we observe a fourfold gain explained by BIF's downward contribution system: Laferrière freezes its children post-initialization while BIF continuously refines subtask representations from every parent observation. This child-level performance is further exemplified in the Child RO score with BIF able to obtain the highest single point reward compared to the baseline models (94.25)

The 3D tasks tests scalability under compositional coupling. In Synthetic 3D, BIF nearly doubles AUC over Vanilla GPBO (BIF UCB: 14,986 vs. 7,678) and achieves a 58\% gain over Laferrière (9,473), only being surpassed by ADD GP. With respect to parent $R^2$, BIF UCB is within a standard error of the top performing model while being over four times that of Vanilla. BIF PI's child $R^2$ is the top scorer, outperforming the next best model ADD GP by 39\% and an eightfold improvement over Laferrière while the top child RO score is achieved by BIF through the EI acquisition function. By Corollary \ref{main_cor:hierarchical_regret}, reducing child kernel mismatch through continuous downward feedback directly tightens the cumulative regret ceiling — these empirical gains have a formal guarantee. BIF reaches peak optimization within just 60 queries (Fig. \ref{fig:main_text_plots}), confirming that bidirectional decomposition of aggregate feedback into per-child updates effectively mitigates the curse of dimensionality. With respect to the Levy 3D task, BIF EI is the best model at finding the single best point for both parent and children. However, can only achieve the second best score in the rest of the tested metrics (ADD GP is best in parent and child $R^2$ and AUC).

Finally, the Hartmann optimization environment is meant to show BIF's performance with exceptionally high number of children. This is due to BIF's expected degradation in performance as the number of children increases since the process of partitioning information in the downward information flow mechanism becomes noisier as different child functions add noise. In this regime, BIF is unable to get the top performance in any metric, however it is still able to achieve the second best score in parent RO and child $R^2$ with UCB while child RO and AUC perform better with BIF PI.

\paragraph{Real-World Applications (Table \ref{table:performance_real}).}

On the neurostimulation benchmark, defined by high observation noise and complex spatiotemporal interactions, BIF achieves the strongest reconstruction fidelity at every level of the hierarchy. Parent $R^2$ reaches 81.43 with the EI acquisition function, exceeding Vanilla GPBO by 8.4 points and more than doubling Laferrière, while BIF UCB's child $R^2$ of 63.98 represents a 3.6× improvement over Laferrière. This child fidelity translates over to optimization performance, with BIF UCB achieving a perfect child RO score. BIF UCB also achieves the highest AUC, confirming sustained sample efficiency throughout training. Vanilla GPBO and Laferrière achieve higher RO in this configuration; sensitivity analysis (Appendix \ref{sec-complementary:opt_target} and \ref{app:hp_selection}) reveals this as a deliberate, controllable tradeoff governed by hyperparameter $\kappa$ — the reported configuration maximizes AUC, prioritizing structural robustness and clinical interpretability essential in \textit{in vivo} settings where subtask misspecification carries direct safety consequences.

BIF's dominance is most pronounced on HPO. Across GAN and Lasso tasks, BIF variants achieve optimal scores in seven of eight tracked metrics. The sole exception — parent $R^2$ on HPO-Lasso, where Deep GP scores 89.27 versus BIF PI's 79.96 — is neutralized by a six-fold runtime overhead (Table \ref{table:wall_clock}), making Deep GP impractical in time-limited regimes. BIF PI delivers the highest AUC on HPO-Lasso and the best child $R^2$ by a margin of 30\% over the next best ADD GP. Most notably, BIF UCB achieves top RO (96.50), parent $R^2$ (12.19), and BIF PI achieves top child $R^2$ on HPO-GAN despite an ANOVA analysis confirming 54.5\% of GAN hyperparameters exhibit non-additive interactions — well outside BIF's quasi-linear assumption — demonstrating that bidirectional feedback and prior map aggregation extracts reliable optimization signals despite structural misspecification.

\paragraph{Modularity and Subtask Transfer.}
\label{sec:modularity_results}

BIF's modular architecture enables pre-trained child GPs to be transplanted into novel composite objectives, bypassing cold-start initialization. As shown in Figure~\ref{fig:modularity}, modular initialization yields immediate gains in early-stage parent performance over training from scratch (Appendix~\ref{app:modularity}). End-to-end BIF eventually surpasses modular initialization in child $R^2$, a crossover theoretically predicted by Corollary~\ref{main_cor:hierarchical_regret}: mismatched child priors temporarily inflate the regret ceiling until the downward contribution pathway re-aligns the hierarchy to the target task. This confirms both the practical value of modular transfer for data-scarce settings and the corrective necessity of bidirectional flow.

\paragraph{Ablation Study.}
\label{sec:ablation_study}
To isolate the contribution of each information pathway, we remove the upward (prior maps) and downward (response contribution) modules independently across three benchmarks (Table \ref{tab:ablation}, further details in Appendix \ref{sec-complementary:ablation}). The two pathways impose distinct costs when ablated, reflecting their mechanistically distinct roles. Removing the upward pathway primarily degrades AUC — by 14\% on HPO-Lasso and 6\% on Michalewicz — because the parent loses structured early-stage guidance and must rediscover informative regions from scratch, slowing convergence without permanently capping performance. Removing the downward pathway is far more destructive to hierarchy fidelity: child $R^2$ collapses by 58\% on HPO-Lasso, as sub-models cannot refine their representations beyond initialization. Crucially, this child collapse does not proportionally damage parent RO — the parent GP learns to dynamically discount stagnant priors — but it directly degrades the kernel quality bound established in Corollary \ref{main_cor:hierarchical_regret}, explaining why AUC suffers even when the final optimum is recovered. Full BIF outperforms both ablated variants in 11 of 12 metrics, confirming that state-of-the-art performance requires joint integration of both pathways.    

\section{Conclusion}

We have presented Bidirectional Information Flow (BIF), a hierarchical GP-BO framework that establishes continuous two-way communication between parent and child surrogates via upward prior map aggregation and a downward softmax credit-assignment scheme. Together, these mechanisms allow the full hierarchy to learn in parallel from a single stream of environment queries, overcoming the information bottlenecks inherent in unidirectional H-GP methods.

Theoretically, we prove that GP-UCB cumulative regret scales linearly with kernel mismatch (Thm. \ref{thm:main_theorem}) and, in the hierarchical case, is bounded by the sum of child-level errors in linearly composed settings (Cor. \ref{main_cor:hierarchical_regret}) — formally justifying continuous child refinement. Proposition \ref{prop:prior_map} strengthens the connection to BIF by proving that the upward prior map converges uniformly to the true objective at a rate governed by child prediction error and residual uncertainty, directly accelerating parent-level convergence as children improve. Proposition \ref{prop:contribution_bias} proves that BIF's downward softmax contribution introduces a Lipschitz-bounded bias that vanishes exactly when child contributions are equal, providing a formal convergence guarantee for the credit-assignment pathway. Together, these results elevate BIF from a heuristically motivated architecture to one with end-to-end proven guarantees.

Empirically, BIF delivers consistent performance across all tested domains: it places first or top-two in 38/40 tests, an eightfold child $R^2$ improvement over \citet{laferriere_hierarchical_2020} on the 3D synthetic task, and optimal scores on nine of ten metrics across real-world HPO benchmarks with 54.5\% non-additive interactions, well outside BIF's quasi-linear assumption. On the neurostimulation benchmark, BIF achieves the highest parent $R^2$, child RO, child $R^2$, and AUC of any method, directly supporting its adoption in safety-critical and data scarce optimization protocols. 

Naturally, limitations remain. The quasi-linear additive assumption introduces theoretical degradation at high nonlinearity, affecting both the quality of the prior maps and the accuracy of the inferred data points in the downward pathway, which is studied in Appendix \ref{app:nonlinearity}. This means that our theoretical insights do not support the performance of our model outside of the linear regime despite its empirical performance demonstrating otherwise. Secondly, BIF's performance is expected to drop as substantially more child dimensions are added as observed in BIF not being able to get the best score in the Hartmann 6D problem. This is likely caused by inaccuracies introduced in the decomposition method introduced since with more children present, there is more noise injected into the inferred data points of the other children. In particular, the decomposition is identifiable up to additive constants, which can be removed through centering or an equivalent shift convention. Additionally, the use of exact GP inference scales cubically with observations. Sparse inducing-point methods and amortized surrogates such as Prior-Data Fitted Networks are the most promising paths forward. For now, BIF's combination of sample efficiency, modularity, and theoretical grounding makes it a strong candidate wherever the optimization target decomposes into interacting subtasks and every query carries a cost — most urgently, in neurostimulation and related safety-critical domains. 
    
\paragraph{Acknowledgments}

This research was supported by the New Frontiers in Research Fund Exploration (NFRFE-2022-00394) and Transformation (NFRFT-2024-00209), the Natural Sciences and Engineering Research Council of Canada (NSERC; RGPIN-2023-04370), and the TransMedTech Institute (Projet-0131-AI-DBS). This research was enabled in part by compute resources provided by Mila (mila.quebec) and the Digital Research Alliance of Canada (alliancecan.ca). Juan David Guerra acknowledges funding support from NSERC and FRQNT. Marco Bonizzato is inventor of a patent (CA3,146,323) for a device enabling cortical stimulation during movement. He is also shareholder in a start-up company developing invasive neurostimulation technologies based on this patent. Guillaume Lajoie acknowledges support from the NSERC Discovery Grant (RGPIN-2018-04821), the Canada Research Chair in Neural Computations and Interfacing, and a Canada-CIFAR AI Chair. Numa Dancause received a Discovery Grant (RGPIN-2022-04210) from the Natural Sciences and Engineering Research Council of Canada (NSERC) to ND. 

\bibliographystyle{plainnat}

\bibliography{references}

@misc{surjanovic_levy_2013,
	title = {Levy {Function}},
	url = {https://www.sfu.ca/~ssurjano/levy.html},
	urldate = {2026-10-05},
	author = {Surjanovic, Sonja and Bingham, Derek},
	year = {2013},
}

@misc{surjanovic_holder_2013,
	title = {Holder {Table} {Function}},
	url = {https://www.sfu.ca/~ssurjano/holder.html},
	urldate = {2026-10-05},
	author = {Surjanovic, Sonja and Bingham, Derek},
	year = {2013},
}

@misc{surjanovic_hartmann_2013,
	title = {Hartmann 6-{Dimensional} {Function}},
	url = {https://www.sfu.ca/~ssurjano/hart6.html},
	urldate = {2026-10-05},
	author = {Surjanovic, Sonja and Bingham, Derek},
	year = {2013},
}

@misc{xiao_fashion-mnist_2017,
	title = {Fashion-{MNIST}: a {Novel} {Image} {Dataset} for {Benchmarking} {Machine} {Learning} {Algorithms}},
	shorttitle = {Fashion-{MNIST}},
	url = {http://arxiv.org/abs/1708.07747},
	doi = {10.48550/arXiv.1708.07747},
	urldate = {2026-04-16},
	publisher = {arXiv},
	author = {Xiao, Han and Rasul, Kashif and Vollgraf, Roland},
	month = sep,
	year = {2017},
	note = {arXiv:1708.07747 [cs]},
}

@article{tibshirani_regression_1996,
	title = {Regression {Shrinkage} and {Selection} {Via} the {Lasso}},
	volume = {58},
	issn = {0035-9246},
	url = {https://doi.org/10.1111/j.2517-6161.1996.tb02080.x},
	doi = {10.1111/j.2517-6161.1996.tb02080.x},
	number = {1},
	urldate = {2026-04-13},
	journal = {Journal of the Royal Statistical Society: Series B (Methodological)},
	author = {Tibshirani, Robert},
	month = jan,
	year = {1996},
	pages = {267--288},
}

@misc{surjanovic_rosenbrock_2013,
	title = {Rosenbrock {Function}},
	url = {https://www.sfu.ca/~ssurjano/rosen.html},
	urldate = {2026-04-13},
	journal = {Virtual Library of Simulation Experiments: Test Functions and Datasets},
	author = {Surjanovic, Sonja and Bingham, Derek},
	year = {2013},
}

@misc{surjanovic_michalewicz_2013,
	title = {Michalewicz {Function}},
	url = {https://www.sfu.ca/~ssurjano/michal.html},
	urldate = {2026-04-13},
	journal = {Virtual Library of Simulation Experiments: Test Functions and Datasets},
	author = {Surjanovic, Sonja and Bingham, Derek},
	year = {2013},
}

@misc{duvenaud_additive_2011,
	title = {Additive {Gaussian} {Processes}},
	url = {http://arxiv.org/abs/1112.4394},
	doi = {10.48550/arXiv.1112.4394},
	urldate = {2026-04-13},
	publisher = {arXiv},
	author = {Duvenaud, David and Nickisch, Hannes and Rasmussen, Carl Edward},
	month = dec,
	year = {2011},
	note = {arXiv:1112.4394 [stat]},
}

@misc{kingma_adam_2017,
	title = {Adam: {A} {Method} for {Stochastic} {Optimization}},
	shorttitle = {Adam},
	url = {http://arxiv.org/abs/1412.6980},
	doi = {10.48550/arXiv.1412.6980},
	urldate = {2026-01-26},
	publisher = {arXiv},
	author = {Kingma, Diederik P. and Ba, Jimmy},
	month = jan,
	year = {2017},
	note = {arXiv:1412.6980 [cs]},
}

@book{rasmussen_gaussian_2005,
	title = {Gaussian {Processes} for {Machine} {Learning}},
	isbn = {978-0-262-25683-4},
	url = {https://doi.org/10.7551/mitpress/3206.001.0001},
	doi = {10.7551/mitpress/3206.001.0001},
	urldate = {2026-05-01},
	publisher = {The MIT Press},
	author = {Rasmussen, Carl Edward and Williams, Christopher K. I.},
	month = nov,
	year = {2005},
}

@inproceedings{abbasi-yadkori_improved_2011,
	title = {Improved {Algorithms} for {Linear} {Stochastic} {Bandits}},
	volume = {24},
	url = {https://papers.nips.cc/paper_files/paper/2011/hash/e1d5be1c7f2f456670de3d53c7b54f4a-Abstract.html},
	urldate = {2025-09-30},
	booktitle = {Advances in {Neural} {Information} {Processing} {Systems}},
	publisher = {Curran Associates, Inc.},
	author = {Abbasi-yadkori, Yasin and Pál, Dávid and Szepesvári, Csaba},
	year = {2011},
}

@misc{vakili_information_2021,
	title = {On {Information} {Gain} and {Regret} {Bounds} in {Gaussian} {Process} {Bandits}},
	url = {http://arxiv.org/abs/2009.06966},
	doi = {10.48550/arXiv.2009.06966},
	urldate = {2025-09-24},
	publisher = {arXiv},
	author = {Vakili, Sattar and Khezeli, Kia and Picheny, Victor},
	month = mar,
	year = {2021},
	note = {arXiv:2009.06966 [stat]},
}

@article{pedregosa_scikit-learn_2011,
	title = {Scikit-learn: {Machine} {Learning} in {Python}},
	volume = {12},
	issn = {1533-7928},
	shorttitle = {Scikit-learn},
	url = {http://jmlr.org/papers/v12/pedregosa11a.html},
	number = {85},
	urldate = {2025-09-10},
	journal = {Journal of Machine Learning Research},
	author = {Pedregosa, Fabian and Varoquaux, Gaël and Gramfort, Alexandre and Michel, Vincent and Thirion, Bertrand and Grisel, Olivier and Blondel, Mathieu and Prettenhofer, Peter and Weiss, Ron and Dubourg, Vincent and Vanderplas, Jake and Passos, Alexandre and Cournapeau, David and Brucher, Matthieu and Perrot, Matthieu and Duchesnay, Édouard},
	year = {2011},
	pages = {2825--2830},
}

@misc{bogunovic_misspecified_2021,
	title = {Misspecified {Gaussian} {Process} {Bandit} {Optimization}},
	url = {http://arxiv.org/abs/2111.05008},
	doi = {10.48550/arXiv.2111.05008},
	urldate = {2025-08-13},
	publisher = {arXiv},
	author = {Bogunovic, Ilija and Krause, Andreas},
	month = nov,
	year = {2021},
	note = {arXiv:2111.05008 [cs]},
}

@misc{chew_bilbo_2025,
	title = {{BILBO}: {BILevel} {Bayesian} {Optimization}},
	shorttitle = {{BILBO}},
	url = {http://arxiv.org/abs/2502.02121},
	doi = {10.48550/arXiv.2502.02121},
	urldate = {2025-08-11},
	publisher = {arXiv},
	author = {Chew, Ruth Wan Theng and Nguyen, Quoc Phong and Low, Bryan Kian Hsiang},
	month = may,
	year = {2025},
	note = {arXiv:2502.02121 [cs]},
}

@article{srinivas_gaussian_2012,
	title = {Gaussian {Process} {Optimization} in the {Bandit} {Setting}: {No} {Regret} and {Experimental} {Design}},
	volume = {58},
	issn = {0018-9448, 1557-9654},
	shorttitle = {Gaussian {Process} {Optimization} in the {Bandit} {Setting}},
	url = {http://arxiv.org/abs/0912.3995},
	doi = {10.1109/TIT.2011.2182033},
	number = {5},
	urldate = {2025-08-07},
	journal = {IEEE Transactions on Information Theory},
	author = {Srinivas, Niranjan and Krause, Andreas and Kakade, Sham M. and Seeger, Matthias},
	month = may,
	year = {2012},
	note = {arXiv:0912.3995 [cs]},
	pages = {3250--3265},
}

@misc{laferriere_samlafhierarchical-gaussian-process_2019,
	title = {samlaf/hierarchical-gaussian-process},
	url = {https://github.com/samlaf/hierarchical-gaussian-process},
	urldate = {2025-05-13},
	author = {Laferriere, Samuel},
	month = jul,
	year = {2019},
	note = {original-date: 2019-07-06T19:14:22Z},
}

@misc{tighineanu_transfer_2022,
	title = {Transfer {Learning} with {Gaussian} {Processes} for {Bayesian} {Optimization}},
	url = {http://arxiv.org/abs/2111.11223},
	doi = {10.48550/arXiv.2111.11223},
	urldate = {2025-05-12},
	publisher = {arXiv},
	author = {Tighineanu, Petru and Skubch, Kathrin and Baireuther, Paul and Reiss, Attila and Berkenkamp, Felix and Vinogradska, Julia},
	month = mar,
	year = {2022},
	note = {arXiv:2111.11223 [stat]},
}

@article{smith_optimizing_1991,
	title = {Optimizing aircraft performance with adaptive, integrated flight/propulsion control},
	volume = {113},
	url = {https://ntrs.nasa.gov/citations/19910039021},
	urldate = {2025-05-11},
	journal = {ASME, Transactions, Journal of Engineering for Gas Turbines and Power},
	author = {Smith, R. H. and Chisholm, J. D. and Stewart, J. F.},
	month = jan,
	year = {1991},
	note = {NTRS Author Affiliations: McDonnell Aircraft Co., NASA Flight Research Center
NTRS Report/Patent Number: ASME PAPER 90-GT-252
NTRS Document ID: 19910039021
NTRS Research Center: Legacy CDMS (CDMS)},
}

@article{bonizzato_autonomous_2023,
	title = {Autonomous optimization of neuroprosthetic stimulation parameters that drive the motor cortex and spinal cord outputs in rats and monkeys},
	volume = {4},
	issn = {2666-3791},
	url = {https://www.sciencedirect.com/science/article/pii/S2666379123001180},
	doi = {10.1016/j.xcrm.2023.101008},
	number = {4},
	urldate = {2025-05-05},
	journal = {Cell Reports Medicine},
	author = {Bonizzato, Marco and Guay Hottin, Rose and Côté, Sandrine L. and Massai, Elena and Choinière, Léo and Macar, Uzay and Laferrière, Samuel and Sirpal, Parikshat and Quessy, Stephan and Lajoie, Guillaume and Martinez, Marina and Dancause, Numa},
	month = apr,
	year = {2023},
	pages = {101008},
}

@article{auer_finite-time_2002,
	title = {Finite-time {Analysis} of the {Multiarmed} {Bandit} {Problem}},
	volume = {47},
	issn = {1573-0565},
	url = {https://doi.org/10.1023/A:1013689704352},
	doi = {10.1023/A:1013689704352},
	language = {en},
	number = {2},
	urldate = {2025-05-04},
	journal = {Machine Learning},
	author = {Auer, Peter and Cesa-Bianchi, Nicolò and Fischer, Paul},
	month = may,
	year = {2002},
	pages = {235--256},
}

@misc{gardner_gpytorch_2021,
	title = {{GPyTorch}: {Blackbox} {Matrix}-{Matrix} {Gaussian} {Process} {Inference} with {GPU} {Acceleration}},
	shorttitle = {{GPyTorch}},
	url = {http://arxiv.org/abs/1809.11165},
	doi = {10.48550/arXiv.1809.11165},
	urldate = {2025-05-04},
	publisher = {arXiv},
	author = {Gardner, Jacob R. and Pleiss, Geoff and Bindel, David and Weinberger, Kilian Q. and Wilson, Andrew Gordon},
	month = jun,
	year = {2021},
	note = {arXiv:1809.11165 [cs]},
}

@inproceedings{damianou_deep_2013,
	title = {Deep {Gaussian} {Processes}},
	issn = {1938-7228},
	url = {https://proceedings.mlr.press/v31/damianou13a.html},
	language = {en},
	urldate = {2025-04-23},
	booktitle = {Proceedings of the {Sixteenth} {International} {Conference} on {Artificial} {Intelligence} and {Statistics}},
	publisher = {PMLR},
	author = {Damianou, Andreas and Lawrence, Neil D.},
	month = apr,
	year = {2013},
	pages = {207--215},
}

@inproceedings{fox_hierarchical_2007,
	title = {Hierarchical {Dirichlet} processes for tracking maneuvering targets},
	url = {https://ieeexplore.ieee.org/document/4408155},
	doi = {10.1109/ICIF.2007.4408155},
	urldate = {2025-04-07},
	booktitle = {2007 10th {International} {Conference} on {Information} {Fusion}},
	author = {Fox, Emily B. and Sudderth, Erik B. and Willsky, Alan S.},
	month = jul,
	year = {2007},
	pages = {1--8},
}

@inproceedings{fyshe_hierarchical_2012,
	title = {Hierarchical {Latent} {Dictionaries} for {Models} of {Brain} {Activation}},
	issn = {1938-7228},
	url = {https://proceedings.mlr.press/v22/fyshe12.html},
	language = {en},
	urldate = {2025-04-07},
	booktitle = {Proceedings of the {Fifteenth} {International} {Conference} on {Artificial} {Intelligence} and {Statistics}},
	publisher = {PMLR},
	author = {Fyshe, Alona and Fox, Emily and Dunson, David and Mitchell, Tom},
	month = mar,
	year = {2012},
	pages = {409--421},
}

@article{ruberg_application_2023,
	title = {Application of {Bayesian} approaches in drug development: starting a virtuous cycle},
	volume = {22},
	copyright = {2023 Springer Nature Limited},
	issn = {1474-1784},
	shorttitle = {Application of {Bayesian} approaches in drug development},
	url = {https://www.nature.com/articles/s41573-023-00638-0},
	doi = {10.1038/s41573-023-00638-0},
	language = {en},
	number = {3},
	urldate = {2024-08-12},
	journal = {Nature Reviews Drug Discovery},
	publisher = {Nature Publishing Group},
	author = {Ruberg, Stephen J. and Beckers, Francois and Hemmings, Rob and Honig, Peter and Irony, Telba and LaVange, Lisa and Lieberman, Grazyna and Mayne, James and Moscicki, Richard},
	month = mar,
	year = {2023},
	pages = {235--250},
}

@article{laferriere_hierarchical_2020,
	title = {Hierarchical {Bayesian} {Optimization} of {Spatiotemporal} {Neurostimulations} for {Targeted} {Motor} {Outputs}},
	volume = {28},
	issn = {1558-0210},
	url = {https://ieeexplore.ieee.org/abstract/document/9062604?casa_token=3F0As8DHhk0AAAAA:xe_5rOHkQiGtRsrwZ04f_Mz80gL0l7myA94LkISY6yAfow2QnJrRrYchrxNO-drr98gUDAZKeYo},
	doi = {10.1109/TNSRE.2020.2987001},
	number = {6},
	urldate = {2024-08-12},
	journal = {IEEE Transactions on Neural Systems and Rehabilitation Engineering},
	author = {Laferrière, Samuel and Bonizzato, Marco and Côté, Sandrine L. and Dancause, Numa and Lajoie, Guillaume},
	month = jun,
	year = {2020},
	note = {Conference Name: IEEE Transactions on Neural Systems and Rehabilitation Engineering},
	pages = {1452--1460},
}

@article{hensman_hierarchical_2013,
	title = {Hierarchical {Bayesian} modelling of gene expression time series across irregularly sampled replicates and clusters},
	volume = {14},
	issn = {1471-2105},
	url = {https://doi.org/10.1186/1471-2105-14-252},
	doi = {10.1186/1471-2105-14-252},
	number = {1},
	urldate = {2024-06-27},
	journal = {BMC Bioinformatics},
	author = {Hensman, James and Lawrence, Neil D. and Rattray, Magnus},
	month = aug,
	year = {2013},
	pages = {252},
}

@inproceedings{bonilla_multi-task_2007,
	title = {Multi-task {Gaussian} {Process} {Prediction}},
	volume = {20},
	url = {https://papers.nips.cc/paper_files/paper/2007/hash/66368270ffd51418ec58bd793f2d9b1b-Abstract.html},
	urldate = {2024-06-25},
	booktitle = {Advances in {Neural} {Information} {Processing} {Systems}},
	publisher = {Curran Associates, Inc.},
	author = {Bonilla, Edwin V and Chai, Kian and Williams, Christopher},
	year = {2007},
}

@article{an_hierarchical_2021,
	title = {Hierarchical deep learning models using transfer learning for disease detection and classification based on small number of medical images},
	volume = {11},
	copyright = {2021 The Author(s)},
	issn = {2045-2322},
	url = {https://www.nature.com/articles/s41598-021-83503-7},
	doi = {10.1038/s41598-021-83503-7},
	language = {en},
	number = {1},
	urldate = {2024-06-07},
	journal = {Scientific Reports},
	publisher = {Nature Publishing Group},
	author = {An, Guangzhou and Akiba, Masahiro and Omodaka, Kazuko and Nakazawa, Toru and Yokota, Hideo},
	month = mar,
	year = {2021},
	pages = {4250},
}

@inproceedings{lawrence_hierarchical_2007,
	address = {New York, NY, USA},
	series = {{ICML} '07},
	title = {Hierarchical {Gaussian} process latent variable models},
	isbn = {978-1-59593-793-3},
	url = {https://doi.org/10.1145/1273496.1273557},
	doi = {10.1145/1273496.1273557},
	urldate = {2024-06-06},
	booktitle = {Proceedings of the 24th international conference on {Machine} learning},
	publisher = {Association for Computing Machinery},
	author = {Lawrence, Neil D. and Moore, Andrew J.},
	month = jun,
	year = {2007},
	pages = {481--488},
}

\onecolumn
\appendix

\section{Gaussian Process based Bayesian Optimization}
\label{app:GPBO}
In the search for an efficient machine learning method to solve optimization tasks, few options are able to outperform Gaussian Processes (GP) based Bayesian Optimization (BO) in the low data regime. Bayesian Optimization is an iterative optimization algorithm where surrogate models predict which next query will provide the most information to the model based on the set of previous environment queries. BO provides access to uncertainty predictions on top of the expected reward from the different actions in the space, allowing for the application of these algorithms into high risk scenarios where knowledge of the potential errors in a prediction is paramount. BO aims to optimize an unknown optimization function $f: X \rightarrow \mathbb{R}$ where after each iteration, a new evaluation of the optimization function $y_i = f(x_i)$ is added to the training dataset $D_n=\{x_i, y_i\}_{i=0}^n$. 

Calculation of these estimates is performed by the surrogate model, which we select to be Gaussian Processes due to its excellence in low data regimes with Upper Confidence Bounds as the acquisition function (refer to Section \ref{sec:queries} for further details). GPs provide the prediction $p(f(x)|x_n,D_n)$ where $f(x) \sim GP(\mu(x), k(x,x'))$ with $\mu(x)$ being the mean of the GP for point $x$ and $k(x,x')$ being the covariance function used to calculate how related two points $(x,x')$ are to each other. To predict the next test point using the GP, we can calculate the mean and variance using the learned noise $\sigma^2_{lik}$ as follows:

\begin{equation}
\label{eq:gpbo_mean_variance}
    \begin{split}
            \mu(x) &= \textbf{k}_x^T(K + \sigma^2_{lik}I)^{-1}\textbf{y} \\
        \sigma(x)^2&=k(x,x) - \textbf{k}_x^T(K + \sigma^2_{lik}I)^{-1}\textbf{k}_x
    \end{split}
\end{equation}

such that $K$ represents the $n\mathbf{x}n$ covariance matrix and $\textbf{k}_x$ represent the $1\mathbf{x}n$ covariance vector within $K$ corresponding to the point $x$. One of the most common covariance functions used in GPs, and the one selected for the children in this work, is the Matern kernel formally defined below:
\begin{equation}
    k_s(x,x') = \sigma^2\frac{2^{1-\nu}}{\Gamma(\nu)}(\sqrt{2\nu}\frac{d}{\rho})^\nu K_\nu(\sqrt{2\nu}\frac{d}{\rho})
\end{equation} such that $\sigma^2$ is the signal variance, $\rho$ the lengthscale, $\nu$ the smoothness parameter, $d$ the Euclidean distance $\|x-x'\|_2$, and $K_\nu$ the modified Bessel function. A thorough guide on Gaussian Processes can be accessed through \citet{rasmussen_gaussian_2005}.

\section{Upper Confidence Bounds}
\label{app:UCB}

In Bayesian Optimization, the acquisition function guides the search for the optimal query location by quantifying the potential value of evaluating each candidate point in the input space. We adopt the \textit{Upper Confidence Bound (UCB)} as our acquisition function due to its principled trade-off between exploration and exploitation, particularly in data-scarce settings.

\begin{definition}[Upper Confidence Bound]
\label{def:ucb}
Given a surrogate Gaussian Process model with posterior predictive mean $\mu(x)$, predictive standard deviation $\sigma(x)$, and query count $n(x)$ for input $x$, the UCB score is computed as:

\begin{equation}
\label{equation:UCB}
    \text{UCB}(x) = \mu(x) + \kappa \cdot \frac{\sigma(x)}{\sqrt{n(x)}}
\end{equation}
where:
\begin{itemize}
    \item $\mu(x)$ represents the surrogate model’s current estimate of the objective at location $x$ (exploitation),
    \item $\sigma(x)$ captures the model’s predictive uncertainty at $x$ (exploration),
    \item $n(x)$ denotes the number of times location $x$ has been queried,
    \item $\kappa \in \mathbb{R}^+$ is a tunable hyperparameter controlling the exploration–exploitation balance.
\end{itemize}
\end{definition}

The UCB acquisition function encourages the model to explore uncertain regions (where $\sigma(x)$ is high) early in training, while progressively prioritizing exploitation as information accumulates and $n(x)$ increases. This dynamic makes UCB especially suitable for online learning systems like BIF, where efficient sample usage is critical.

In BIF, UCB maps are generated independently by each child model over its respective subspace. These maps are then aggregated into a prior map for the parent model (Section \ref{sec:upward_flow}) and reused in the downward contribution mechanism (Section \ref{sec:downward_flow}), thereby ensuring consistent acquisition strategies across the entire hierarchy.

\section{Hardware Setup}
\label{app:hardware_setup}
The models were run without the use of GPU acceleration and implemented using the GPyTorch Libary \citep{gardner_gpytorch_2021}. Approximate runtime is 1.5 query/s with a maximum budget of 100 queries on Intel 12th Gen i7 processor with 32 Gb of memory.

\section{Compute Efficiency}
\label{app:compute_efficiency}
\begin{table}[ht]
\caption{Empirical wall clock speeds of all tested models. Optimization done over 100 queries on the Michalewicz dataset. Times given as minutes : seconds.}
\label{table:wall_clock}
\centering
\begin{tabular}{cccccc}
\hline
           & Vanilla & Laferrière & BIF & Deep GP & ADD GP-UCB \\ \hline
Clock Time & 0:07             & 0:57                & 1:14         & 5:11             & 0:15 
\\ \hline
\end{tabular}
\end{table}

Runtime is a first-class criterion for surrogate model selection, particularly in resource-constrained or embedded settings where strict inference deadlines apply. To complement BIF's theoretical $\mathcal{O}(n^3)$  complexity, we conducted an empirical wall-clock study benchmarking all frameworks under the Michalewicz synthetic function. Results are reported in Table \ref{table:wall_clock}.

As expected, the Vanilla GP achieves the lowest wall-clock time (7 s total), having no hierarchical structure to maintain — it computes the posterior predictive distribution directly at each step. This simplicity, however, comes at the cost of expressiveness. Introducing a parent GP to coordinate sub-GPs incurs non-trivial overhead, as evidenced by the gap between ADD GP-UCB — which runs independent child models without a shared parent — and the Laferrière and BIF architectures, both of which maintain a parent GP with structural dependencies over child outputs. Since BIF and Laferrière impose comparable computational loads on their upward-pass mechanisms (both transmit prior means to the parent GP), the runtime differential between them isolates the incremental cost of BIF's downward information flow module, which amounts to only 17 seconds over the full 100-query trajectory — a modest overhead for the expressiveness gained. Notably, BIF is not the most expensive model in the suite; the Deep GP incurs approximately 4× the runtime of BIF, a consequence of its deeply nested kernel evaluations.

We argue that BIF's runtime profile is highly favorable given its empirical performance gains. Per-iteration, BIF requires approximately 0.74 s of training time. In scientific applications with expensive physical acquisition — such as deep brain stimulation (DBS) parameter tuning, where each data point may require weeks of clinical assessment — sub-second per-query inference is effectively negligible. Moreover, BIF's hierarchical structure is inherently amenable to parallelization: child GPs are conditionally independent given the parent and can be evaluated concurrently before propagating information upward. Although native parallelization was not implemented in this study, it constitutes a straightforward engineering extension with the potential for substantial wall-clock reductions in large-scale deployments.

\section{Hyperparameter Testing and Selection}
\label{app:hp_selection}

\begin{figure}
    \centering
    \includegraphics[width=1\linewidth]{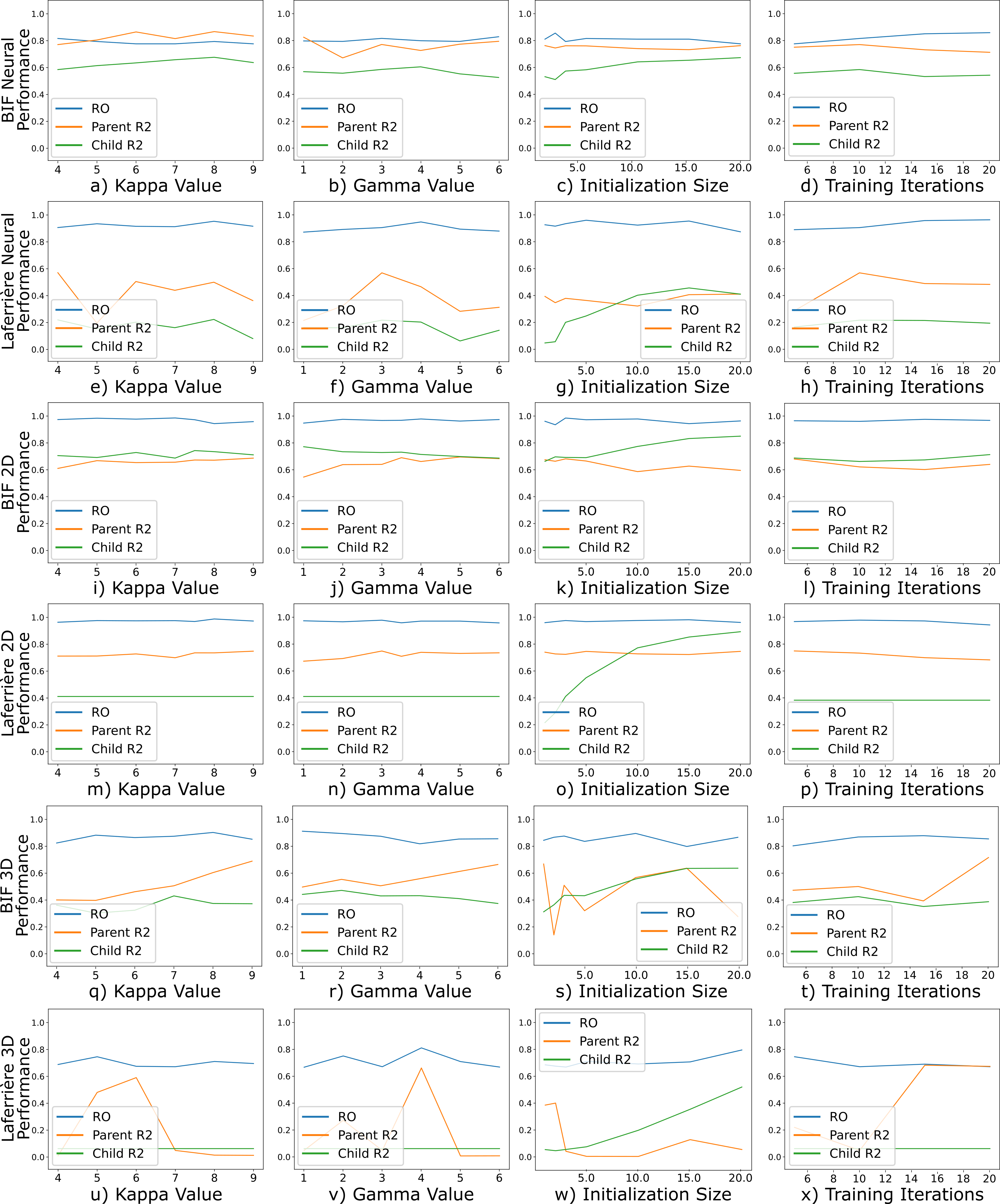}
    \caption{Hyper Parameter Search Results for Max Seen Normalization. Due to time constraints and the number of hyperparameters, search is run one by one for each parameter. Baseline parameters are kappa 7.5, gamma 3, initialization set size of 3, and 10 training iterations. (a-d) represent hyperparameter search for BIF in the Neural Dataset and (e-h) for Laferrière on the neural dataset. (i-l) shows the hyperparameter search for BIF and (m-p) for Laferrière on the 2D synthetic dataset. Finally, (q-t) present the results for BIF on the 3D dataset while (u-x) present Laferrièree}
    \label{fig:hp_search_neural}
\end{figure}

\begin{table}[]
\caption{Optimal Hyperparameters for all models, datasets, and normalization types. Note that 2D and 3D datasets refer to the synthetic 2D and synthetic 3D datasets.}
\label{table:optimal_hp}
\vspace{2mm}
\centering
\begin{tabular}{cccccccccc}
                            &                          & \multicolumn{8}{c}{Normalization}                                                 \\ \cline{3-10} 
                            &                          & \multicolumn{4}{c}{Z-Score}             & \multicolumn{4}{c}{Maximum Observed Division}   \\ \hline
Dataset                     & Hyper Parameter          & 2D & 3D & Neural & Subtract & 2D & 3D & Neural & Subtract \\ \hline
\multirow{4}{*}{BIF}        & Kappa $\kappa$           & 9        & 8        & 8      & 7.5      & 4        & 7        & 7      & 8        \\
                            & Gamma $\gamma$           & 4        & 5        & 4      & 2        & 2        & 2        & 1      & 5        \\
                            & Child Initialization $r$ & 3        & 2        & 1      & 1        & 5        & 3        & 2      & 1        \\
                            & Training Iterations $t$  & 10       & 20       & 15     & 10       & 5        & 20       & 5      & 20       \\ \hline
\multirow{4}{*}{Laferriere} & Kappa $\kappa$           & 7.5      & 7        & 7      & 8        & 7.5      & 5        & 8      & 8        \\
                            & Gamma $\gamma$           & 1        & 3        & 4      & 6        & 6        & 4        & 5      & 5        \\
                            & Child Initialization $r$ & 1        & 3        & 3      & 1        & 1        & 1        & 5      & 1        \\
                            & Training Iterations $t$  & 10       & 5        & 10     & 10       & 5        & 15       & 10     & 15       \\ \hline
\multirow{2}{*}{Vanilla}    & Kappa $\kappa$           & 9        & 6        & 8      & 9        & 7.5      & 6        & 5      & 4        \\
                            & Training Iterations $t$  & 20       & 5        & 15     & 20       & 5        & 15       & 5      & 20      
\end{tabular}
\end{table}
In Figure \ref{fig:hp_search_neural} we show the hyperparameter search for the BIF and Laferrière models in all datasets for the max seen response normalization technique. The same hyperparameter search is done for the z-scoring normalization technique as well as the vanilla model. Balancing these hyperparameters, we selected a kappa value of 4.0, gamma of 3, initialization set size of 6, and 10 training iterations.

\section{Parent Gaussian Process Environment Visualization}
\label{app:parent_heatmap}
\begin{figure}
    \centering
    \includegraphics[width=1\linewidth]{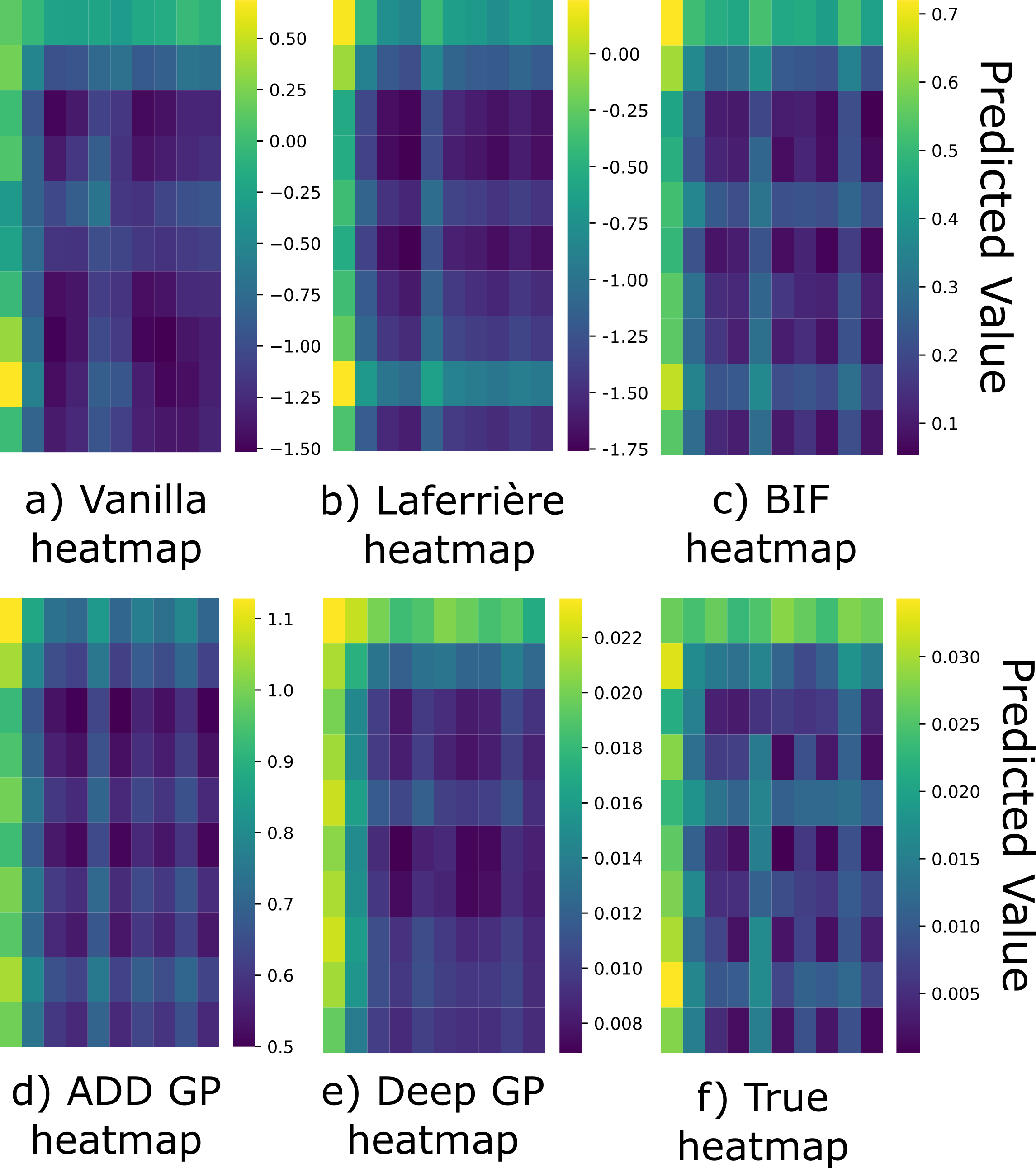}
    \caption{Parent Environment Reconstructions for Neural Dataset (a-d). True heatmap denotes the ground truth and all other heatmaps correspond to a specific model type. The more the heatmaps resemble the ground truth, the higher the $R^2$ score.}
    \label{fig:parent_heatmaps}
\end{figure}
In this section, we present the environment reconstruction of the parent processes as done for the children in Figure \ref{fig:main_text_plots}. Note we cannot display the heatmaps of the Synthetic 3D task as this would require a 4D heatmap visualization. Due to this, we strictly rely on the $R^2$ scores of the parent as a guide for environment reconstruction accuracy. We create these state maps as a way to visualize the $R^2$ scores presented in the performance figures. All models have the same method applied to them, where they are queried over the entire state space and results are plotted in the 2D heatmap presented in Figure \ref{fig:parent_heatmaps}. The true heatmap (d,h) represent the ground truth of the environment, setting a baseline of what the other maps should resemble. For the Neural Stimulation dataset, we see Vanilla GPBO and \citet{laferriere_hierarchical_2020} able to select the optimal stimulation locations. However, for the rest of the map, the BIF model is able to reconstruct the state space more accurately than the other models.

\begin{table}[h]
\centering
\resizebox{\textwidth}{!}{%
\begin{tabular}{lccccc}
\toprule
 & \textbf{Parent RO} & \textbf{Parent R$^2$} & \textbf{Average Child RO} & \textbf{Average Child R$^2$} & \textbf{AUC} \\
\midrule
Neural Stimulation & & $\uparrow$ & $\uparrow$ & $\uparrow$ & $\uparrow$ \\
Michalewicz        & $\downarrow$ & $\uparrow$ & $\uparrow$ & $\uparrow$ & $\uparrow$ \\
HPO - GAN          & & & & & $\uparrow$ \\
HPO - Lasso        & & $\downarrow$ & $\uparrow$ & & $\uparrow$ \\
Synthetic 3D       & $\uparrow$ & & & & \\
Holder 2D          & & & & $\uparrow$ & \\
Levy 3D            & & $\downarrow$ & & $\downarrow$ & $\downarrow$ \\
Hartmann 6D        & & $\downarrow$ & & & \\
\bottomrule
\end{tabular}
}
\caption{Paired t-test across 30 repetitions significance difference testing. Results with $\uparrow$ indicate significantly higher scores for BIF against the next best method, $\downarrow$ indicate significantly lower scores for BIF, and no arrow denotes no significant difference}
\label{tab:significance}
\end{table}

\section{Significance Testing}
\label{app:significance}
To better understand the performance of our models compared to the state-of-the-art methods presented in this work, we ran an additional experiment where the results of all the models following 30 identical and random seeds were put through a paired t-test. The results of this analysis are presented in Table \ref{tab:significance}. It is important to note that these results were run on random seeds and therefore may have slightly different outcomes than those presented in the maintext. With these results, it is clear that BIF is exceptional in maximizing the AUC, where the BIF model family is significantly higher in 4/8 datasets including all 3 real-world applications. This metric in particular is incredibly important to optimization problems as achieving a high AUC score can only be done by both achieving high scores in all metrics and doing so quickly. Similarly, the BIF model family achieves significantly higher child RO scores on 3 of the tested landscapes, with no model being able to outperform it significantly. However, BIF is not purely better than all tested models. In the parent $R^2$ metric, BIF performs significantly worse than the state-of-the-art in 3/8 datasets. Despite this, the table shows us that BIF in general is as good if not better than any other of the tested models as it achieves significantly higher scores than the baselines in 13 different scenarios.

\section{Bidirectional Information Flow Modularity}
\label{app:modularity}
\begin{figure}
    \centering
    \includegraphics[width=1\linewidth]{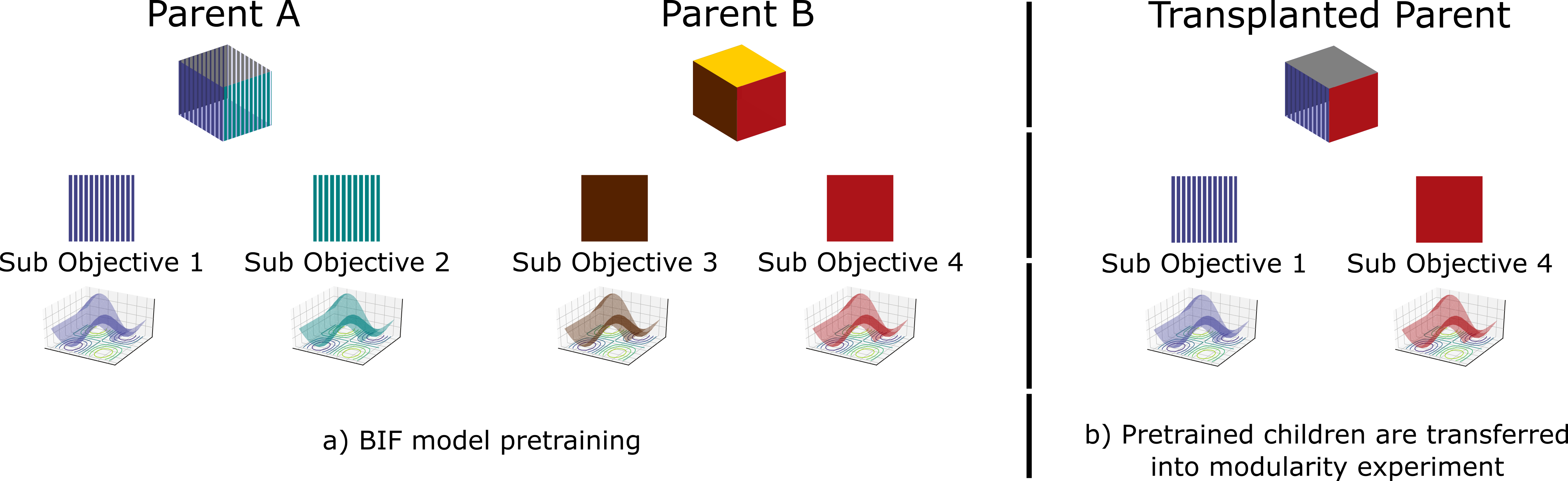}
    \caption{Visual Diagram for Modularity Experiment. BIF hierarchy are trained on initial datasets with their children being placed into a new parent to boost performance.}
    \label{fig:modularity_sketch}
\end{figure}
\begin{figure}
    \centering
    \includegraphics[width=1\linewidth]{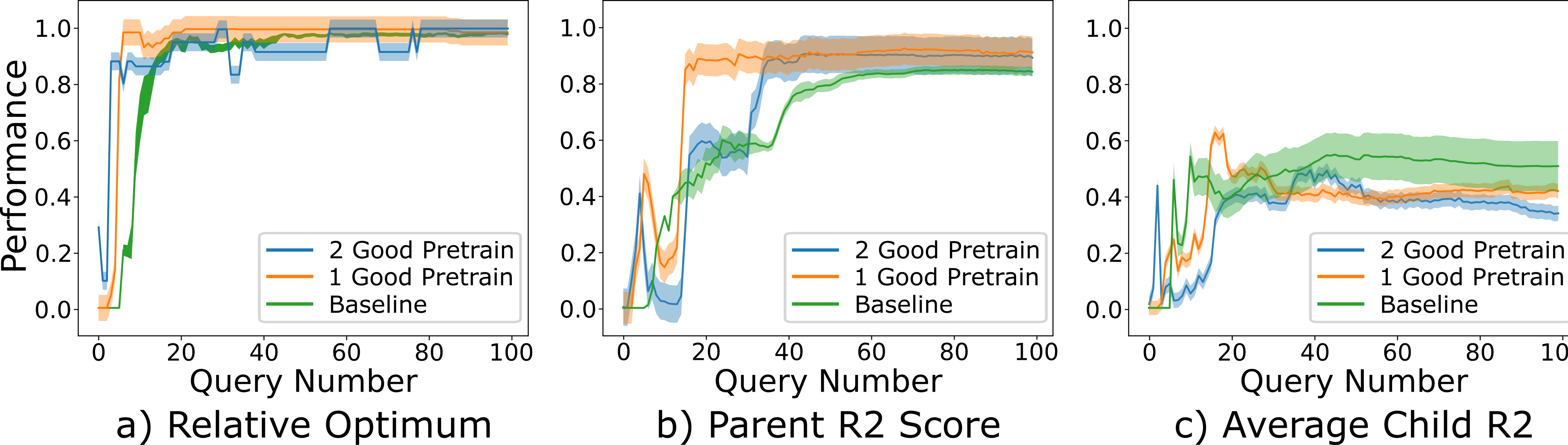}
    \caption{Effect of swapping pretrained children on BIF performance. Good pretraining denotes switching to the same child task and bad denotes using the pretrained child on a different task. Baseline performance is regular BIF training as presented in Algorithm \ref{alg:BIF}.}
    \label{fig:modularity}
\end{figure}

Here we provide a more in depth discussion and interpretation of the modularity experiments. As Stated in Section \ref{sec:modularity}, these experiments are designed to show one of the key benefits of using a modular structure - the segmentation of tasks. In the BIF setup, we aim to minimize the need of pretraining the children while still providing boosts to the overall optimization goal. In this modularity experiment, the initialization set outlined in the model setup is replaced by inserting children who have been trained previously on a different overarching task while erasing the prior queries, preserving only the hyperparameters. This experiment can be thought of as form of transfer learning, where pretrained modules (i.e. children) are sent to novel tasks to boost performance. Previous works have looked at how hierarchical GPs can efficiently do this type of task \citep{tighineanu_transfer_2022}. We define a set of experiments as the optimization of three different tasks:
\begin{enumerate}
    \item \(f(x) = -(x-2)^2 +2, \space g(y) = -\frac{x-1}{2}^4 + 2, \) and parent function \(h(x,y) = \frac{f(x)+g(y)}{(x-y)^2 + \epsilon}\);
    \item \(f'(x) = \frac{sin(2\pi x)}{x-1} + 2,  \space g'(y) = \frac{sin(2\pi x)}{x-2} + 2, \) and parent function \( h'(x,y) = \frac{f(x)+g(y)}{(x-y)^2 + \epsilon}\);
    \item the modular task \(h''(x,y) = \frac{f(x) + g'(y)}{(x-y)^2 + 2 + \epsilon}\)
\end{enumerate}
Note that the modular task is made of the child functions from the two previous tasks, allowing the transfer of knowledge between parent optimization tasks. This mimics real-world use of information, where doing a hierarchical task can reuse pretrained information that was previously used to build a different hierarchy. A diagram of this process is presented in Figure \ref{fig:modularity_sketch}

We observe in Figure \ref{fig:modularity} that for early training steps, the pretraining regimen works in boosting the performance of the parent model early and maintains better performance for the Relative Optimum and the $R^2$ of the parent. However, the performance of the reconstruction of the children is hindered, experiencing a slightly lower performance than the baseline model. We expect the eventual higher performance of the baseline model compared to the properly pretrained models could be due to information from the inserted children containing what can be considered noise from the previous optimization tasks. However, future work could investigate the effect of this noise as the children are continuously swapped into more optimization tasks. This could help the GP find the true function from the noise and we expect a significantly higher performance boost.
\section{Robustness Testing}
\label{app:robustness}

In this section, we present a set of experiments aimed at pushing the BIF algorithm to its limits to show its robustness to different scenarios it may face when used in real world applications.

\subsection{Bidirectional Information Flow Nonlinearity Robustness}
\label{app:nonlinearity}

\begin{figure}
    \centering
    \includegraphics[width=1\linewidth]{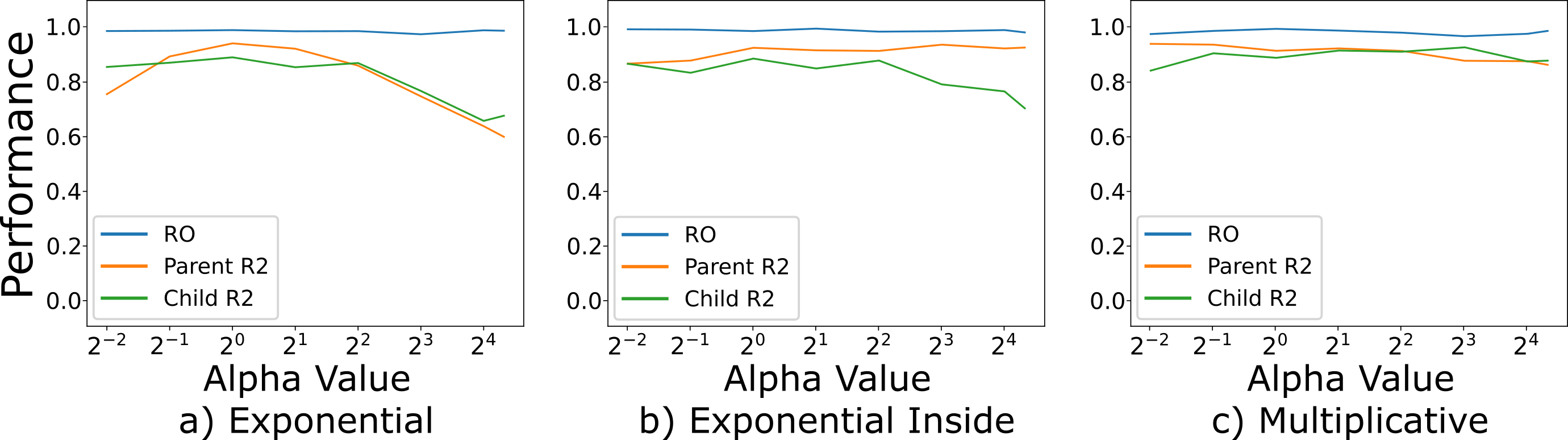}
    \caption{Results of \textbf{Nonlinearity Experiments} described in Appendix \ref{app:nonlinearity}. We aim to show the inherent biases of the model due to the assumption of linear interactions between submodels. Higher $\alpha$ values signify an increase in nonlinearity as shown in the description of the experiments in Appendix \ref{app:nonlinearity}. The alpha values scale exponentially with a base of 2.}
    \label{fig:nonlinearity}
\end{figure}

\begin{table}[t]
\caption{Performance on Subtractive interaction task. Note that Vanilla GPBO denotes a non-hierarchical GPBO method consequently removing its ability to represent the child $R^2$ task. Performance metrics are described in Section \ref{sec:performance_metrics}}
\label{table:subtractive}
\vspace{2mm}
\centering
\begin{tabular}{cccccc}
\hline
\textbf{Dataset}             & \textbf{Model}                                                           & \textbf{Relative Optimum} & \textbf{Parent $R^2$} & \textbf{Child $R^2$} & \textbf{AUC}       \\ \hline
\multirow{3}{*}{Subtraction} 
    & Vanilla GPBO                                                             & \textbf{99.72}       & \textbf{96.91}     & -                 & 14,110.62           \\
    & Laferrière                                     & 98.31       & 73.17     & 7.62             & 15,467.07          \\
    & BIF & 99.16       & 93.25     & \textbf{81.65}    & \textbf{22,650.28} \\ \hline
\end{tabular}

\end{table}

Machine learning models are typically constructed with a certain task in mind, leading their architecture to be designed for the given task. In BIF, the target task is neurostimulation optimization, which combines interactions in a predominantly linear manner \citep{bonizzato_autonomous_2023, laferriere_hierarchical_2020}. To elucidate the limitations of our model, we present experiments aimed at targeting the weaknesses of the BIF model. As stated in the methodologies section (Section \ref{sec:additive_assumption}), the composition of the prior maps and the contribution system assume an intrinsic quasi-linear additive relationship between the children when being combined to the parent task. With this assumption, we aim to see to what end of nonlinearity the BIF model can still learn efficiently before its inductive biases break down. We propose 4 forms of nonlinearity in order to understand better what domains the model is robust to:

\begin{itemize}
    \item Exponential Nonlinearity: \(h(x,y) = [f(x) + g(y)]^\alpha\)
    \item Exponential Inside Nonlinearity: \(h(x,y) = f(x)^\alpha + g(y)\)
    \item Multiplicative Factor Nonlinearity: \(h(x,y) = f(x) + g(y) + \alpha f(x)g(y)\)
    \item Subtraction: \((h(x,y) = f(x) - g(y)\)
\end{itemize}

$\alpha$ scales the degree of nonlinearity. We set $f(x) = \sin(x)$ and $g(y) = \tanh(y)$ as simple child objectives to highlight how nonlinearity affects learning through the structural biases of BIF.

Here we show how BIF is able to generalize to child interactions lying outside of its expected domain. Figure \ref{fig:nonlinearity} illustrates that as $\alpha$ scales, a metric to denote the degree of nonlinearity, the effects are negligible with respect to the relative optimum across all tests, and in the specific case of the multiplicative nonlinearity test, increasing the alpha value has no significant effects on any of the metrics (Fig. \ref{fig:nonlinearity}c). In the two other nonlinearity experiments, Figures \ref{fig:nonlinearity}(a-b), we see a more significant effect on the model. In the exponential experiment (Fig. \ref{fig:nonlinearity}a), the model has a 37\% reduction in the parent's $R^2$ score and a 24\% reduction of the average child's $R^2$ score. In the exponential inside experiment, only the average child $R^2$ experiences a significant drop of 17\% (Fig. \ref{fig:nonlinearity}b), aligning with the fact that it is in essence a simplified version of the exponential experiment. The disproportionate effect of nonlinearity on the children is expected, as the contribution system from which the children train from is the most heavily reliant on the linearity assumption while the other metrics, particularly the parent, is only taking the linearity assumption as a prior to the mean. In light of this, the model is still highly robust to nonlinearity, as only high alpha values are able to affect the model's performance. We also see robustness to subtractive relationships, directly opposing the kernel construction, in Table \ref{table:subtractive}. The BIF model attains its highest performance across all metrics and datasets in this subtraction dataset, showing that the kernel construction being negative instead of positive has little to no effect on the performance of the model, demonstrating its ability to learn in this regime.

\subsection{Effects of Noise on Bidirectional Information Flow}
\label{app:noise}

\begin{figure}
    \centering
    \includegraphics[width=1\linewidth]{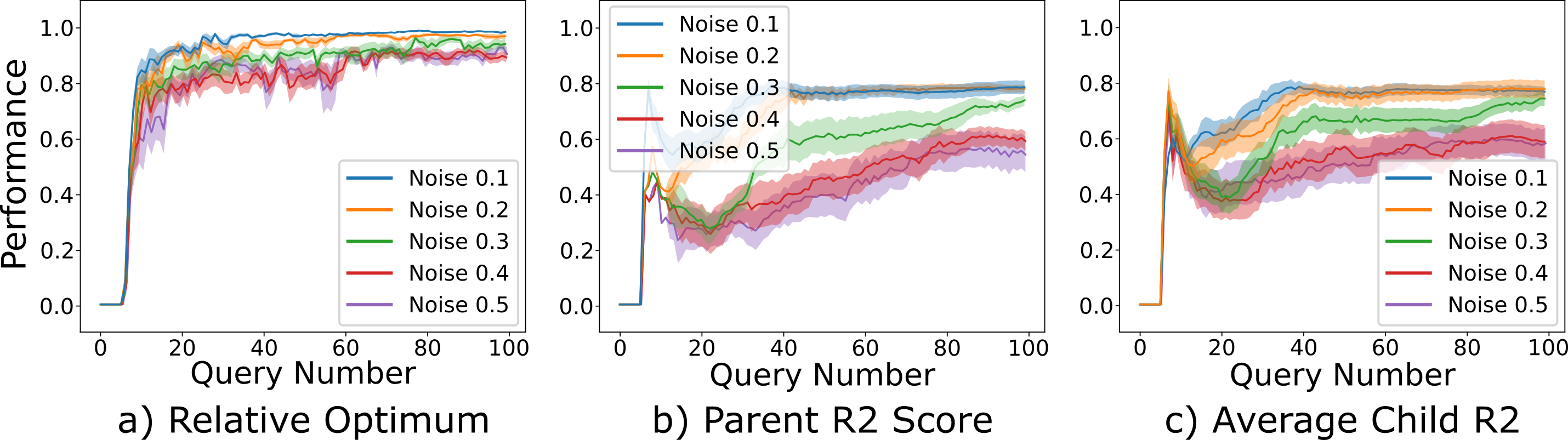}
    \caption{Performance of \textbf{Bidirectional Information Flow with Scaling Noise} as detailed in Appendix \ref{app:noise}. Higher values of noise mean higher values of $\beta$ in the description of the experiment in Appendix \ref{app:noise}. Relative Optimum is more affected than other metrics with scaling noise, followed by average coefficient of determination for the children, and parent coefficient of determination not significantly affected.}
    \label{fig:noise}
\end{figure}

Robustness to noise is an important factor when deploying machine learning models to the real world. Typically, a model will be able to learn through the noise when sufficient data points are present. However, in our target application of low data regimes with high noise, such as neurostimulation, our model must be able to deal with this task. Here, we investigate the model's performance compared to baseline models as more noise is introduced to the data. This is implemented by adding a noise scaling factor where at each query to the environment, noise is sampled from a Normal distribution and added to the environment response such that $y_{noisy} = y + \epsilon$ where $ \epsilon \sim N(0, \beta)$, $y$ denotes the ground truth output, and $\beta$ is a parameter that adjusts the amount of noise present in the dataset. With this setup, we scale $\beta$ to see how robust the model is as noise increases.

In Figure \ref{fig:noise}, we present the results of scaling noise. Most notable of the noise testing results is in Figure \ref{fig:noise}b where the parent $R^2$ score is seemingly unaffected by the increased noise while the RO score and child $R^2$ have diminishing scores as noise increases. Interestingly, the child $R^2$ score looks to have the same amount of information learned as noise increases but with different starting points with higher noise lowering the initial performance. Running a statistical test on this observation shows the average gain in performance is 0.23 points with a standard deviation of 0.017. This suggests that the amount of information learned remains constant with minimal variability as noise is scaled, demonstrating that the BIF model is, in part, robust to noise with respect to the $R^2$ of the children. Finally, the RO score of the BIF model shown in Figure \ref{app:noise} is affected most significantly by scaling noise with all instances starting at nearly the same RO score and diverging throughout training with higher noise values resulting in lower RO scores.

\begin{figure}
    \centering
    \includegraphics[width=0.7\linewidth]{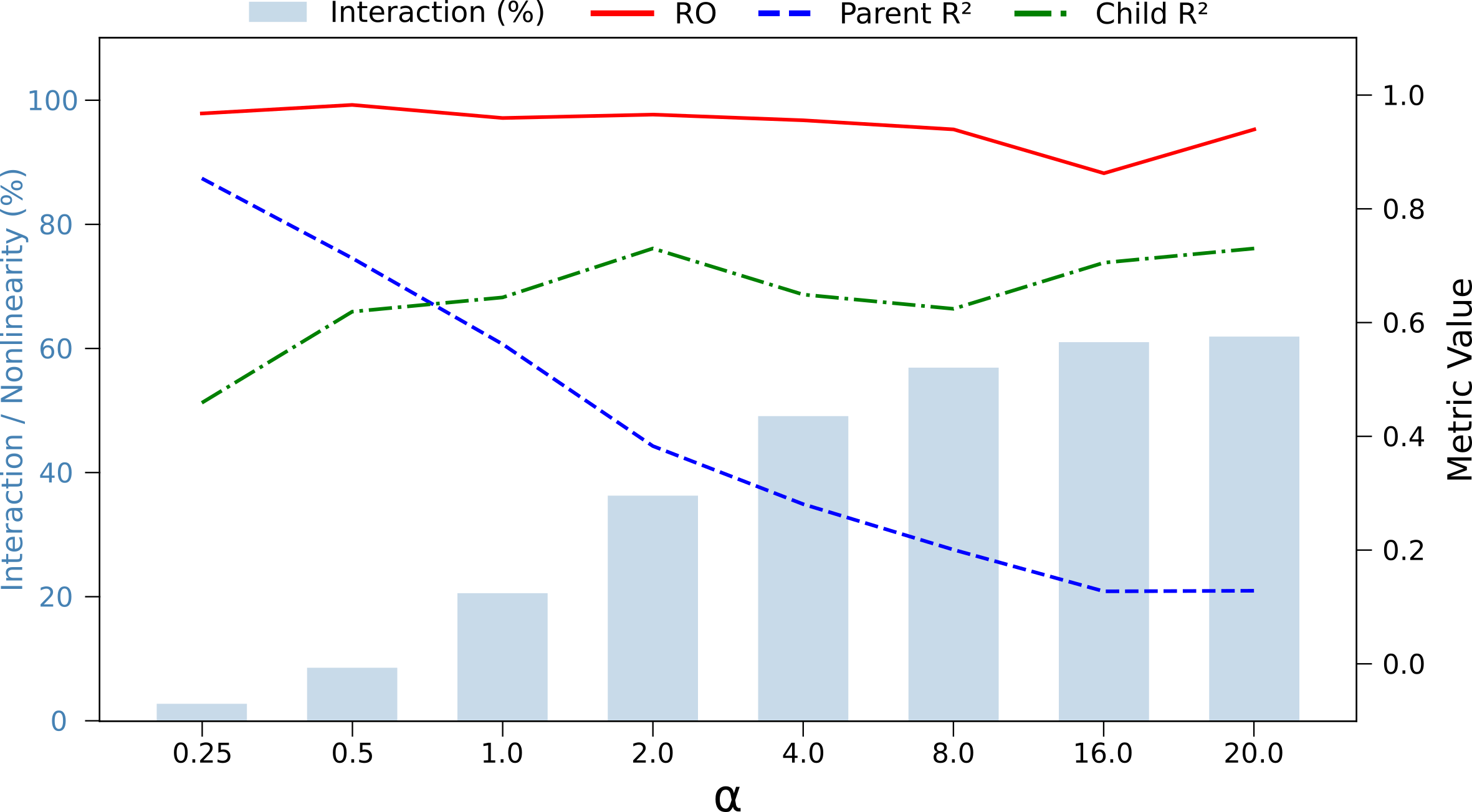}
    \caption{Performance of BIF - UCB  on Rosenbrock nonlinearity dataset. $\alpha$ acts as a scaling parameter, where higher values increase the degree of nonlinear interactions shown in the bar graph. Line plots show BIF performance as nonlinearity increases.}
    \label{fig:rosenbrock}
\end{figure}

\subsection{Rosenbrock Nonlinearity Testing}
\label{sec-complementary:rosenbrock_nonlinearity_testing}

We further analyze BIF's capacity to maintain robustness against escalating nonlinear interactions using the Rosenbrock function. Results are visualized in Figure \ref{fig:rosenbrock}. As $\alpha$ increases, the dataset's non-additivity scales drastically from 3\% to 64\%. Theoretically, this should violently disrupt BIF’s quasi-linear additive assumption, leading to widespread performance degradation. Nevertheless, BIF's RO remains nearly unaffected, maintaining a score above 86\% across all nonlinearity thresholds. An interesting dynamic emerges in the $R^2$ metrics: as nonlinear interactions dominate the space, the child $R^2$ improves while the parent $R^2$ degrades at a proportional rate.

These results fortify the robustness tests previously detailed in Section \ref{app:nonlinearity}, demonstrating that BIF gracefully handles adversarial environments where its core assumptions are violated.

\section{Mechanistic Insights}
\label{app:mechanistic_insights}

\begin{table}[ht]
\caption{Ablation study on the upward and downward information flow modules presented in BIF. No Up Module signifies the removal of the prior maps from children to parent. No Down Module removes the contribution system continually training the children post-initialization.}
\label{tab:ablation}
\centering
\begin{tabular}{llccc}
\toprule
 & & No Up Module & No Down Module & Full BIF \\
\midrule
\multirow{4}{*}{Sub 2D}
 & RO       & $98.43 \pm 0.29$ & $98.67 \pm 0.34$ & $\mathbf{99.16 \pm 0.22}$ \\
 & Parent $R^2$ & $92.63 \pm 1.08$ & $92.79 \pm 1.62$ & $\mathbf{93.25 \pm 0.63}$ \\
 & Child $R^2$  & $74.76 \pm 2.55$ & $10.99 \pm 1.67$ & $\mathbf{81.65 \pm 1.94}$ \\
 & AUC      & $20,251.86$ & $13,813.10$ & $\mathbf{22,650.28}$ \\
\midrule
\multirow{4}{*}{Michalewicz}
 & RO       & $98.91 \pm 0.60$ & $\mathbf{99.89 \pm 0.06}$ & $98.87 \pm 0.60$ \\
 & Parent $R^2$ & $78.37 \pm 3.71$ & $79.24 \pm 1.16$ & $\mathbf{86.10 \pm 0.88}$ \\
 & Child $R^2$  & $47.43 \pm 4.38$ & $32.08 \pm 4.74$ & $\mathbf{61.88 \pm 3.25}$ \\
 & AUC      & $18,079.62$ & $16,782.96$ & $\mathbf{18,587.90}$ \\
\midrule
\multirow{4}{*}{HPO - Lasso}
 & RO       & $99.85 \pm 0.11$ & $99.92 \pm 0.02$ & $\mathbf{99.98 \pm 0.02}$ \\
 & Parent $R^2$ & $68.635 \pm 2.98$ & $69.00 \pm 2.29$ & $\mathbf{75.35 \pm 2.84}$ \\
 & Child $R^2$  & $37.68 \pm 4.44$ & $21.57 \pm 3.94$ & $\mathbf{50.96 \pm 5.79}$ \\
 & AUC      & $16,549.95$ & $16,708.63$ & $\mathbf{17,906.14}$ \\
\bottomrule
\end{tabular}
\end{table}

\subsection{BIF Module Ablation Study}
\label{sec-complementary:ablation}

As the name Bidirectional Information Flow implies, BIF relies on two distinct computational streams: one propagating information upward (child to parent) and one downward (parent to child). To isolate their differential impacts, an ablation study was conducted. This ablation study was conducted by removing either upward or downward information flow pathways and comparing them with the full BIF model (Table \ref{tab:ablation}).

Removing the upward pathway (prior maps from children to parent) predictably degrades the Area Under the Curve (AUC). Deprived of early environmental priors, the overarching model requires more steps to converge, even if it eventually reaches a comparable RO. Conversely, removing the downward pathway (the response contribution system) causes child $R^2$ scores to stagnate, as the sub-models cannot refine their representations post-initialization. Crucially, without the downward module, child $R^2$ drops by up to 85\% in the Sub 2D dataset and 58\% in the HPO - Lasso dataset. However, because the parent GP learns to dynamically disregard stagnant or misleading priors over time, the overall RO remains relatively stable showing BIF's ability to handle faulty initializations. 

The ablation results confirm that BIF’s state-of-the-art performance stems not from isolated mechanisms, but from the joint integration of both upward and downward pathways. Consequently, the full BIF architecture outperforms its ablated counterparts in 11 out of 14 tested metrics. Furthermore, the variation in results between the different module ablations gives a deeper understanding of the role of the prior information and response attribution systems.

\subsection{Optimization Target Adaptability}
\label{sec-complementary:opt_target}

\begin{figure}
    \centering
    \includegraphics[width=\linewidth]{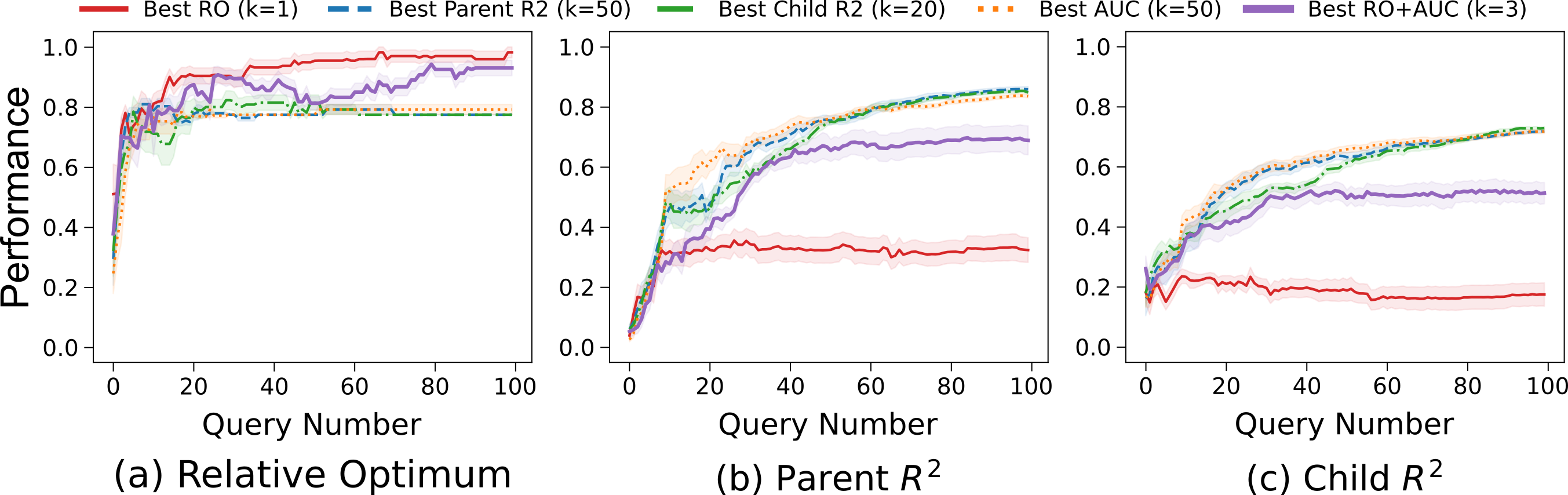}
    \caption{Performance of BIF on Neurostimulation based on different optimization objectives. Different colours denote the optimal setup to attain the desired goal. RO + AUC represents a combined optimization setup that prioritizes an average score of RO and AUC. $\kappa$ values represent the optimal kappa for the optimization criteria}
    \label{fig:HP Tuning with different objectives}
\end{figure}

As stated in Section \ref{app:hp_selection}, BIF’s optimal hyperparameters were originally selected to maximize AUC, specifically highlighting the model's sample efficiency and novel response contribution system. This diverges from standard BO protocols, which typically select models based purely on identifying the single best target point, equivalent to maximizing the RO. Due to this decision, the BIF model sees performance drops in RO, an important metric when considering which BO framework to use. 

To demonstrate BIF's flexibility, we conducted a hyperparameter sweep on the neural dataset where the selection criteria varied depending on the desired optimization goal. The optimization performance of these models is visualized in Figure \ref{fig:HP Tuning with different objectives}. When tuned exclusively for RO, BIF achieves a dominant score of 98.1—surpassing the combined (RO + AUC) tuning by 5 points, and exceeding the originally reported RO in Table \ref{table:performance} by over 16 points. However, optimizing for solely RO collapses performance in parent $R^2$, child $R^2$, and overall AUC. This exposes the inherent bias of single-point optimization while tuning for comprehensive metrics like AUC or $R^2$ yields a vastly more balanced and robust surrogate model.

With these results, we confirm BIF's ability to optimize different BO criteria effectively while further justifying the use of alternative hyperparameter selection criteria. This grants practitioners the confidence to deploy the framework across a variety of distinct contexts where optimization priorities may differ.

\section{Regret Analysis of GP-UCB with a Learned Kernel}
\label{sec:regret_learned}
This section provides a detailed regret analysis for the GP-UCB algorithm when the kernel is learned or misspecified at each time step. We begin by establishing the problem setup and key assumptions, then prove a series of lemmas that bound the perturbation effects of kernel mismatch on the posterior mean and variance. Finally, we combine these results to derive a comprehensive cumulative regret bound.

\subsection{Problem Setup and Notation}

\begin{description}
    \item[Decision Space \& Horizon:] The action set is a compact space $\calX \subset \reals^d$ over a time horizon $T$.
    \item[Objective Function:] The unknown function $f:\calX \to \reals$ is assumed to be a sample from a zero-mean Gaussian Process (GP) with a true, stationary kernel $k^*$, denoted as $f \sim \mathcal{GP}(0, k^*)$.
    \item[Observations:] At each time step $t \in \{1, \dots, T\}$, an algorithm selects a point $\bx_t \in \calX$ and observes a noisy reward $y_t = f(\bx_t) + \varepsilon_t$, where $\varepsilon_t \sim \calN(0, \sigma^2)$ are i.i.d. Gaussian noise terms.
    \item[Algorithm Model:] The algorithm uses a GP with a kernel $k_{t-1}$ which is learned based on data up to time $t-1$. Let $D_{t-1} = \{(\bx_i, y_i)\}_{i=1}^{t-1}$ be the observation history. The posterior distribution over $f(\bx)$ at time $t$ is $\calN(\mu_{t-1}(\bx), \sigma_{t-1}^2(\bx))$, where:
    \begin{align}
        \mu_{t-1}(\bx) &= \bk_{t-1}(\bx)^T (\bK_{t-1} + \sigma^2 \bI)^{-1} \by_{t-1} \\
        \sigma_{t-1}^2(\bx) &= k_{t-1}(\bx, \bx) - \bk_{t-1}(\bx)^T (\bK_{t-1} + \sigma^2 \bI)^{-1} \bk_{t-1}(\bx)
    \end{align}
    Here, $\by_{t-1} = [y_1, \dots, y_{t-1}]^T,(\bK_{t-1})_{ij} = k_{t-1}(\bx_i, \bx_j)\text{, and }\bk_{t-1}(\bx) = [k_{t-1}(\bx_1, \bx), \dots, k_{t-1}(\bx_{t-1}, \bx)]^T$.
    \item[Selection Rule:] The algorithm employs the GP Upper Confidence Bound (UCB) selection rule \cite{srinivas_gaussian_2012}:
    $$ \bx_t = \argmax_{\bx \in \calX} \left( \mu_{t-1}(\bx) + \sqrt{\beta_t} \sigma_{t-1}(\bx) \right) $$
    for some predefined $\beta_t$ value.
    \item[Oracle Model:] For the analysis, we define the ``oracle'' model that uses the true kernel $k^*$ and the same data $D_{t-1}$. It's posterior mean and variance are denoted $\tilde{\mu}_{t-1}(\bx)$ and $\tilde{\sigma}_{t-1}^2(\bx)$ respectively.
    \item[Kernel Mismatch:] The kernel mismatch at step $t$ is defined as the uniform deviation:
    $$ \Delta_t := \sup_{\bx, \bx' \in \calX} |k^*(\bx, \bx') - k_{t-1}(\bx, \bx')| $$
\end{description}

\subsection{Assumptions}

Our analysis relies on the following standard assumptions.

\begin{assumption}[Bounded RKHS Norm]
\label{ass:bounded_rkhs}
The true function $f$ has a bounded norm in the Reproducing Kernel Hilbert Space (RKHS) $\calH_{k^*}$ associated with the true kernel $k^*$. We assume there exists a constant $B > 0$ such that $\norm{f}_{\calH_{k^*}} \leq B$.
\end{assumption}

\begin{assumption}[Uniformly Bounded Kernel Variance]
\label{ass:bounded_kernel_variance}
The diagonal of both the true and learned kernels is uniformly bounded. We assume there exists a constant $\kappa > 0$ such that for all $t \in \{1, \dots, T\}$:
$$ \sup_{\bx \in \calX} k^*(\bx, \bx) \leq \kappa^2 \quad \text{and} \quad \sup_{\bx \in \calX} k_{t-1}(\bx, \bx) \leq \kappa^2 $$
\end{assumption}

\subsection{Foundational Lemmas}

We first establish three lemmas to build the groundwork for our claim. Lemma 1 is a standard result for GP confidence bounds, which we cite. Lemmas 2 and 3 provide novel  proofs for the perturbation of the posterior mean and variance due to kernel mismatch.

\begin{lemma}[GP Confidence Bound with True Kernel]
\label{lemma:GP_confidence}

Let $\delta \in (0,1)$ and set $\beta_t = 2B^2 + 36d\log(t)\log^2(t/\delta)$. With probability at least $1-\delta$, for all $t \geq 1$, the true function $f(\bx)$ lies within the confidence intervals constructed with the true kernel $k^*$:
\begin{equation}
    |f(\bx) - \tilde{\mu}_{t-1}(\bx)| \leq \sqrt{\beta_t} \tilde{\sigma}_{t-1}(\bx), \quad \forall \bx \in \calX
\end{equation}
\end{lemma}

\begin{proof}
This is a standard result in GP bandit literature. The specific form of $\beta_t$ adapts results on the suprema of Gaussian processes to the bandit setting. A full proof can be found in \cite{vakili_information_2021}, which builds upon the foundational work of \cite{srinivas_gaussian_2012}. We State it here without proof.
\end{proof}

\begin{lemma}[Posterior Mean Perturbation]
\label{lemma:posterior_mean_perturbation}
Under Assumptions 1 and 2, with probability at least $1-\delta$, the absolute difference between the learned and oracle posterior means is uniformly bounded for all $\bx \in \calX$ and $t \in \{1, \dots, T\}$ by:
\begin{equation}
    |\mu_{t-1}(\bx) - \tilde{\mu}_{t-1}(\bx)| \leq \delta_{\mu,t}
\end{equation}
where $\delta_{\mu,t} := \frac{\sqrt{t-1}\Delta_t}{\sigma^2} (B\kappa + \sigma\sqrt{t-1}+ \sigma\sqrt{2\log(\frac{2T}{\delta})})(1 + \frac{\kappa(t-1)}{\sigma^2})$.
\end{lemma}

\begin{proof}
Let $\bC_{t-1} = \bK_{t-1} + \sigma^2\bI$ and $\tilde{\bC}_{t-1} = \bK_{t-1}^* + \sigma^2\bI$ where $K,K^*$ are the learned and true kernels respectively.
The difference is $\mu_{t-1}(\bx) - \tilde{\mu}_{t-1}(\bx) = [\bk_{t-1}(\bx)^T \bC_{t-1}^{-1} \by_{t-1}] - [\bk_{t-1}^*(\bx)^T \tilde{\bC}_{t-1}^{-1} \by_{t-1}]$.
By adding 0 and using the triangle inequality:

$\mu_{t-1}(\bx) - \tilde{\mu}_{t-1}(\bx) = [\bk_{t-1}(\bx)^T \bC_{t-1}^{-1} \by_{t-1}] - [\bk_{t-1}^*(\bx)^T \tilde{\bC}_{t-1}^{-1} \by_{t-1}] + \bk_{t-1}^*(\bx)^T {\bC}_{t-1}^{-1} \by_{t-1} - \bk_{t-1}^*(\bx)^T {\bC}_{t-1}^{-1} \by_{t-1}$

$$\mu_{t-1}(\bx) - \tilde{\mu}_{t-1}(\bx) = (\bk_{t-1}(\bx) - \bk_{t-1}^*(\bx))^T \tilde{\bC}_{t-1}^{-1} \by_{t-1} + \bk_{t-1}^*(\bx)^T (\bC_{t-1}^{-1} - \tilde{\bC}_{t-1}^{-1}) \by_{t-1}$$

$$ |\mu_{t-1}(\bx) - \tilde{\mu}_{t-1}(\bx)| \leq |(\bk_{t-1}(\bx) - \bk_{t-1}^*(\bx))^T \tilde{\bC}_{t-1}^{-1} \by_{t-1}| + |\bk_{t-1}^*(\bx)^T (\bC_{t-1}^{-1} - \tilde{\bC}_{t-1}^{-1}) \by_{t-1}| $$
Now let's bound both terms. \\
\textbf{Term 1}: Using the Cauchy-Schwarz inequality, we can say 
\\ $|(\bk_{t-1}(\bx) - \bk_{t-1}^*(\bx))^T \tilde{\bC}_{t-1}^{-1} \by_{t-1}| \leq \norm{\bk_{t-1}(\bx) - \bk_{t-1}^*(\bx)}_2 \norm{\tilde{\bC}_{t-1}^{-1} \by_{t-1}}_2$.\\
Note that $\norm{\bk_{t-1}(\bx) - \bk_{t-1}^*(\bx)}_2^2 = \sum_{i=1}^{t-1} (k_{t-1}(\bx_i, \bx) - k^*(\bx_i, \bx))^2 \leq (t-1)\Delta_t^2$.\\
So, $\norm{\bk_{t-1}(\bx) - \bk_{t-1}^*(\bx)}_2 \leq \sqrt{t-1}\Delta_t$. \\
\textbf{Term 2}: Using the identity $A^{-1} - B^{-1} = A^{-1}(B-A)B^{-1}$, we get
$$|\bk_{t-1}^*(\bx)^T  (\bC^{-1}_{t-1} - \tilde{\bC}^{-1}_{t-1})  \by_{t-1}| = |\bk_{t-1}^*(\bx)^T \bC_{t-1}^{-1} (\tilde{\bC}_{t-1} - \bC_{t-1}) \tilde{\bC}_{t-1}^{-1} \by_{t-1}|$$
$$|\bk_{t-1}^*(\bx)^T  (\bC^{-1}_{t-1} - \tilde{\bC}^{-1}_{t-1})  \by_{t-1}| \leq \norm{\bk_{t-1}^*(\bx)}_2 \norm{\bC_{t-1}^{-1}}_2 \norm{\tilde{\bC}_{t-1} - \bC_{t-1}}_2 \norm{\tilde{\bC}_{t-1}^{-1} \by_{t-1}}_2$$
We have $\norm{\bk_{t-1}^*(\bx)}_2 \leq \sqrt{t-1}\kappa$ by Assumption 2. For $\|C^{-1}_{t-1}\|$, note that $K_{t-1}$ is positive semi-definite by definition. Then the eigenvalues of $K_{t-1}, \lambda(K_{t-1}), \geq 0$ for all $i$. So,
$$K_{t-1}(x)^Tv = \lambda_iv \text{ for all eigenvectors } v \in k_{t-1}$$
$$(K_{t-1}(x)^T + \sigma^2I)v = \lambda_iv + \sigma^2Iv$$
$$(K_{t-1}(x)^T + \sigma^2I)v = (\lambda_i + \sigma^2I)v$$
Since $\lambda_iv \geq 0$, then $\lambda_i(K_{t-1}(x)^T + \sigma^2I) = \lambda_i(C_{t-1}) \geq 0 + \sigma^2 = \sigma^2$. Now, $\|C_{t-1}^{-1}\|$ is the largest eigenvalue in absolute. Now, if a matrix A has eigenvalues $\lambda_i$, its inverse $A^{-1}$ has eigenvalues $1/\lambda_i$. Since $C_{t-1}$ is symmetric and for all $i$, $\lambda_i(C_{t-1}) \geq \sigma^2$, then $\forall i \lambda_i(C_{t-1}^{-1}) \leq \frac{1}{\sigma^2}$. Therefore, $\|C_{t-1}^{-1}\|_2 \leq \frac{1}{\sigma^2}$. This applies to $\tilde{C}_{t-1}^{-1}$ as well so $\|\tilde{C}_{t-1}^{-1}\|_2 \leq \frac{1}{\sigma^2}$.

Next, we look at the norm $$\norm{\tilde{\bC}_{t-1} - \bC_{t-1}}_2 = \norm{\bK_{t-1}^* + \sigma^2\bI - \bK_{t-1} -\sigma^2\bI}_2$$ 
$$\norm{\tilde{\bC}_{t-1} - \bC_{t-1}}_2  \leq \norm{\bK_{t-1}^* - \bK_{t-1}}_2 = \sqrt{\sum_{i,j} |(K^*_{t-1})_{i,j} - (K_{t-1})_{i,j}|^2} \leq \sqrt{(t-1)^2 \Delta_t^2} = (t-1)\Delta_t$$
$$\norm{\tilde{\bC}_{t-1} - \bC_{t-1}}_2  \leq (t-1)\Delta_t$$
\\
Finally, we need to bound $\norm{\by_{t-1}}_2$. We know the observations are $y_i = f(\bx_i) + \varepsilon_i$. By Assumption 1, $|f(\bx)| \leq \norm{f}_{\calH_{k^*}} \sqrt{k^*(\bx, \bx)} \leq B\kappa$. The noise terms $\varepsilon_i$ are i.i.d. $\calN(0, \sigma^2)$. By a standard concentration argument for sub-Gaussian variables \cite{srinivas_gaussian_2012, abbasi-yadkori_improved_2011}, with probability at least $1-\delta$, $\norm{\bepsilon_{t-1}}_2 \leq \sigma\sqrt{t-1} + \sigma\sqrt{2\log(1/\delta)}$. Thus, $\norm{\by_{t-1}}_2 \leq \norm{\mathbf{f}_{t-1}}_2 + \norm{\bepsilon_{t-1}}_2 \leq B\kappa + \sigma\sqrt{t-1} + \sigma\sqrt{2\log(1/\delta)}$.
Applying a union bound over $t=1, \dots, T$ (replacing $\delta$ with $\delta/(2T)$) and combining terms yields the Stated result.
$$|\mu_{t-1}(\bx) - \tilde{\mu}_{t-1}(\bx)| \leq \frac{\sqrt{t-1}\Delta_t}{\sigma^2} (B\kappa + \sigma\sqrt{t-1}+ \sigma\sqrt{2\log(\frac{2T}{\delta})})(1 + \frac{\kappa(t-1)}{\sigma^2})$$
\end{proof}

\begin{lemma}[Posterior Variance Perturbation]
\label{lemma:posterior_variance_perturbation}
Under Assumption 2, for any $\bx \in \calX$, the absolute difference between the learned and oracle posterior variances is bounded by:
$$ |\sigma_{t-1}^2(\bx) - \tilde{\sigma}_{t-1}^2(\bx)| \leq \delta_{\sigma,t} $$
where $\delta_{\sigma,t} := \frac{\kappa(t-1)\Delta_t}{\sigma^2}(\frac{\kappa(t-1)}{\sigma^2} + 2)$.
\end{lemma}

\begin{proof}
We decompose the difference:
$\sigma_{t-1}^2(\bx) - \tilde{\sigma}_{t-1}^2(\bx) = (k_{t-1}(\bx, \bx) - \tilde{k}_{t-1}(\bx, \bx)) - (\bk_{t-1}(\bx)^T \bC_{t-1}^{-1} \bk_{t-1}(\bx) - \tilde{\bk}_{t-1}(\bx)^T \tilde{\bC}_{t-1}^{-1} \tilde{\bk}_{t-1}(\bx))$.
We know from proof \ref{lemma:posterior_mean_perturbation} that $k_{t-1}(\bx, \bx) - \tilde{k}_{t-1}(\bx, \bx) \leq \Delta_t$. We add 0 to the second term and bound it using the triangle inequality:
$$ |\bk_{t-1}^T \bC_{t-1}^{-1} \bk_{t-1} - \tilde{\bk}_{t-1}^T \tilde{\bC}_{t-1}^{-1} \tilde{\bk}_{t-1}| \leq |\bk^T(\bC^{-1} - \tilde{\bC}^{-1})\bk| + |(\bk - \tilde{\bk})^T \tilde{\bC}^{-1}\bk| + |\tilde{\bk}^T \tilde{\bC}^{-1}(\bk - \tilde{\bk})|. $$
$$ \|\bk_{t-1}^T \bC_{t-1}^{-1} \bk_{t-1} - \tilde{\bk}_{t-1}^T \tilde{\bC}_{t-1}^{-1} \tilde{\bk}_{t-1}\| \leq \|\bk^T(\bC^{-1} - \tilde{\bC}^{-1})\bk\| + \|(\bk - \tilde{\bk})^T \tilde{\bC}^{-1}\bk\| + \|\tilde{\bk}^T \tilde{\bC}^{-1}(\bk - \tilde{\bk})\|. $$
We use the results from proof \ref{lemma:posterior_mean_perturbation} $ \norm{\bC^{-1} - \tilde{\bC}^{-1}}_2 \leq \norm{\bC^{-1}}_2 \norm{\tilde{\bC} - \bC}_2 \norm{\tilde{\bC}^{-1}}_2 \leq \frac{(t-1)\Delta_t}{\sigma^4}$, $\norm{\bk}_2, \norm{\tilde{\bk}}_2 \leq \sqrt{t-1}\kappa$, and $\norm{\bk-\tilde{\bk}}_2 \leq \sqrt{t-1}\Delta_t$.\\
Plugging these results into the different subterms we get:
\begin{enumerate}
    \item $\|\bk^T(\bC^{-1} - \tilde{\bC}^{-1})\bk\| \leq (\sqrt{t-1}\kappa)\frac{(t-1)\Delta_t}{\sigma^4}(\sqrt{t-1}\kappa) = (t-1)\kappa^2\frac{(t-1)\Delta_t}{\sigma^4}$
    \item $\|(\bk - \tilde{\bk})^T \tilde{\bC}^{-1}\bk\| \leq \|(\bk - \tilde{\bk})^T \|\|\tilde{\bC}^{-1}\|\|\bk\| \leq (\sqrt{t-1}\kappa)(\frac{1}{\sigma^2})(\sqrt{t-1}\Delta_t) = \kappa \frac{(t-1)\Delta_t}{\sigma^2}$
    \item $\|\tilde{\bk}^T \tilde{\bC}^{-1}(\bk - \tilde{\bk})\| \leq \|(\bk - \tilde{\bk})^T \|\|\tilde{\bC}^{-1}\|\|\bk\| \leq (\sqrt{t-1}\Delta_t)(\frac{1}{\sigma^2})(\sqrt{t-1}\kappa) = \kappa \frac{(t-1)\Delta_t}{\sigma^2}$
\end{enumerate}
Combining all these subterms yields the bound
$$|\sigma_{t-1}^2(\bx) - \tilde{\sigma}_{t-1}^2(\bx)| \leq (t-1)\kappa^2\frac{(t-1)\Delta_t}{\sigma^4} + \kappa \frac{(t-1)\Delta_t}{\sigma^2} + \kappa \frac{(t-1)\Delta_t}{\sigma^2}$$
$$|\sigma_{t-1}^2(\bx) - \tilde{\sigma}_{t-1}^2(\bx)| \leq \frac{\kappa(t-1)\Delta_t}{\sigma^2}(\frac{\kappa(t-1)}{\sigma^2} + 2)$$
\end{proof}

\subsection{Instantaneous Regret Decomposition}
\label{lemma:instantaneous_regret_decomposition}
The instantaneous regret at step $t$ is $r_t = f(\bx_t^*) - f(\bx_t)$, where $\bx_t^* = \argmax_{\bx \in \calX} f(\bx)$ and $x_t$ selected using the UCB rule.
With probability at least $1-2\delta$, Lemma 1 holds for both $f(\bx_t^*)$ and $f(\bx_t)$:
\begin{align*}
    f(\bx_t^*) &\leq \tilde{\mu}_{t-1}(\bx_t^*) + \sqrt{\beta_t} \tilde{\sigma}_{t-1}(\bx_t^*) \\
    f(\bx_t) &\geq \tilde{\mu}_{t-1}(\bx_t) - \sqrt{\beta_t} \tilde{\sigma}_{t-1}(\bx_t)
\end{align*}
Substituting these into the regret expression:
$$ r_t \leq (\tilde{\mu}_{t-1}(\bx_t^*) - \tilde{\mu}_{t-1}(\bx_t)) + \sqrt{\beta_t}(\tilde{\sigma}_{t-1}(\bx_t^*) + \tilde{\sigma}_{t-1}(\bx_t)) $$
By the UCB selection rule, $\mu_{t-1}(\bx_t) + \sqrt{\beta_t} \sigma_{t-1}(\bx_t) \geq \mu_{t-1}(\bx_t^*) + \sqrt{\beta_t} \sigma_{t-1}(\bx_t^*)$.
This implies $\mu_{t-1}(\bx_t^*) - \mu_{t-1}(\bx_t) \leq \sqrt{\beta_t}(\sigma_{t-1}(\bx_t) - \sigma_{t-1}(\bx_t^*))$.
Now, we relate the oracle quantities to the algorithm's quantities using Lemma \ref{lemma:posterior_mean_perturbation}:
$\tilde{\mu}_{t-1}(\bx_t^*) \leq \mu_{t-1}(\bx_t^*) + \delta_{\mu,t}$ and $\tilde{\mu}_{t-1}(\bx_t) \geq \mu_{t-1}(\bx_t) - \delta_{\mu,t}$.
$$ \tilde{\mu}_{t-1}(\bx_t^*) - \tilde{\mu}_{t-1}(\bx_t) \leq \mu_{t-1}(\bx_t^*)  + \delta_{\mu,t} - \mu_{t-1}(\bx_t) + \delta_{\mu,t}$$
$$ \tilde{\mu}_{t-1}(\bx_t^*) - \tilde{\mu}_{t-1}(\bx_t) \leq \sqrt{\beta_t}(\sigma_{t-1}(\bx_t) - \sigma_{t-1}(\bx_t^*)) + 2\delta_{\mu,t} $$
Substituting this back into the regret bound:
$$ r_t \leq (\tilde{\mu}_{t-1}(\bx_t^*) - \tilde{\mu}_{t-1}(\bx_t)) + \sqrt{\beta_t}(\tilde{\sigma}_{t-1}(\bx_t^*) + \tilde{\sigma}_{t-1}(\bx_t)) $$
$$ r_t \leq \sqrt{\beta_t}(\sigma_{t-1}(\bx_t) - \sigma_{t-1}(\bx_t^*)) + 2\delta_{\mu,t} + \sqrt{\beta_t}(\tilde{\sigma}_{t-1}(\bx_t^*) + \tilde{\sigma}_{t-1}(\bx_t)) $$
Now we bound the oracle standard deviations. From Lemma \ref{lemma:posterior_variance_perturbation}, $|\sigma^2 - \tilde{\sigma}^2| \leq \delta_{\sigma,t}$. We now have the property $-\delta_{\sigma,t} \leq \sigma^2 - \tilde{\sigma}^2 \leq \delta_{\sigma,t}$. A first order Taylor expansion gives $\sqrt{a+b} \approx \sqrt{a} + \frac{b}{2\sqrt{a}}$, so we can approximate $\tilde{\sigma}_{t-1}(\bx) \leq \sigma_{t-1}(\bx) + \delta'_{\sigma,t}$ for $\delta'_{\sigma,t} = \frac{\delta}{2\sigma_{t-1}}$.
A simpler path is to use the property $\sqrt{a + b} \leq \sqrt{a} + \sqrt{b}$ in order to produce the bound $\tilde{\sigma}_{t-1}(\bx) \leq \sigma_{t-1}(\bx) + \sqrt{\delta_{\sigma,t}}$ if we assume $\sigma_{t-1}^2(\bx) \geq \delta_{\sigma,t}$, making $\sqrt{\sigma^2_{t-1}(x) + \delta_{\sigma,t}} \leq \sqrt{\sigma^2_{t-1}(x)} + \sqrt{\delta_{\sigma,t}}$, otherwise the bound is loose. Putting this altogether:
\begin{align*}
r_t &\leq \sqrt{\beta_t}(\sigma_{t-1}(\bx_t) - \sigma_{t-1}(\bx_t^*)) + 2\delta_{\mu,t} + \sqrt{\beta_t}(\sigma_{t-1}(\bx_t^*) + \sqrt{\delta}_{\sigma,t} + \sigma_{t-1}(\bx_t) + \sqrt{\delta}_{\sigma,t}) \\
r_t &\leq 2\sqrt{\beta_t} \sigma_{t-1}(\bx_t) + 2\delta_{\mu,t} + 2\sqrt{\beta_t}\sqrt{\delta}_{\sigma,t}
\end{align*}

\subsection{Bounding the Cumulative Regret}
\label{lemma:cumulative_regret}
The cumulative regret is $R_T = \sum_{t=1}^T r_t$. Summing the instantaneous regret:
$$ R_T \leq \sum_{t=1}^T 2\sqrt{\beta_t} \sigma_{t-1}(\bx_t) + \sum_{t=1}^T (2\delta_{\mu,t} + 2\sqrt{\beta_t}\sqrt{\delta}_{\sigma,t}) $$
Let $\beta_T = \max_{t \leq T} \beta_t$. We bound the first sum using Cauchy-Schwarz:
$$ \sum_{t=1}^T 2\sqrt{\beta_t}\sigma_{t-1}(\bx_t) \leq 2\sqrt{\beta_T} \sqrt{\left(\sum_{t=1}^T \sigma_{t-1}^2(\bx_t)\right) \cdot T} = \sqrt{T\beta_T} \sqrt{\sum_{t=1}^T \sigma_{t-1}^2(\bx_t)} $$
The sum of variances is bounded by:
$$ \sum_{t=1}^T \sigma_{t-1}^2(\bx_t) \leq \sum_{t=1}^T (\tilde{\sigma}_{t-1}^2(\bx_t) + \delta_{\sigma,t}) = \left(\sum_{t=1}^T \tilde{\sigma}_{t-1}^2(\bx_t)\right) + \sum_{t=1}^T \delta_{\sigma,t} $$
The sum of oracle posterior variances is bounded by the maximum information gain $\gamma_T(k^*) = \max_{A \subset \calX, |A|=T} \frac{1}{2} \log \detf(\bI + \sigma^{-2}\bK_A^*)$. Specifically, $\sum_{t=1}^T \tilde{\sigma}_{t-1}^2(\bx_t) \leq \frac{2}{\log(1+\sigma^{-2})} \gamma_T(k^*)$. Let $C = \frac{2}{\log(1+\sigma^{-2})}$.
$$ \sum_{t=1}^T \sigma_{t-1}^2(\bx_t) \leq C\gamma_T(k^*) + \sum_{t=1}^T \delta_{\sigma,t} $$
Putting everything together, we get the final bound
\begin{equation}
    R_T \leq 2\sqrt{T\beta_T(C\gamma_T(k^*) + \sum_{t-1}^T\delta_{\sigma,t})}+ \sum_{t=1}^T (2\delta_{\mu,t} + 2\sqrt{\beta_t\delta_{\sigma,t}})
\end{equation}

\subsection{Main Theorem}
\begin{theorem}{Regret Bound for GP-UCB with a Learned Kernel}
\label{thm:main_theorem}

    Under Assumptions 1 and 2, let $\delta \in (0, 1)$ and set $\beta_t$ appropriately (see proof). With probability at least $1 - \delta$, the cumulative regret $R_T$ is bounded by:
$$ R_T \le 2 \sqrt{T \beta_T \left(C\gamma_T(k^*) + \sum_{t=1}^T \delta_{\sigma,t}\right)} + \sum_{t=1}^T \left(2\delta_{\mu,t} + 2\sqrt{\beta_t \delta_{\sigma,t}}\right) $$
where $C = \frac{2}{\log(1+\sigma^{-2})}$, $\beta_T = \max_{t\le T} \beta_t$, and the perturbation bounds $\delta_{\mu,t}$ and $\delta_{\sigma,t}$ depend on the kernel mismatch $\Delta_t$ as shown in lemmas \ref{lemma:posterior_mean_perturbation} and \ref{lemma:posterior_variance_perturbation}.
\end{theorem}{}

\vspace{0.5cm}

\begin{proof}
    
The overall result relies on several probabilistic events holding simultaneously for all time steps $t \in \{1, \dots, T\}$. Let the total failure probability be $\delta$. We use a union bound to ensure all necessary conditions hold with a total probability of at least $1-\delta$.

The key events required are:
\begin{enumerate}
    \item \textbf{E1:} The GP confidence bound from Lemma 1 must hold for all $t \in \{1, \dots, T\}$.
    \item \textbf{E2:} The concentration bounds on the observation vector norms $\|y_{t-1}\|_2$, used in the proof of Lemma 2, must hold for all $t \in \{1, \dots, T\}$.
\end{enumerate}

We allocate a failure probability of $\delta/2$ to each of these sets of events.

For \textbf{E1}, we need the confidence bound to hold for all $T$ time steps. We apply another union bound: for each step $t$, we require the bound to hold with probability $1 - \frac{\delta}{2T}$. To achieve this, we set the confidence parameter in Lemma 1 to be $\delta' = \frac{\delta}{2T}$. This modifies our confidence width parameter to:
$$ \beta_t = 2B^2 + 36d \log(t) \log^2(t/\delta') = 2B^2 + 36d \log(t) \log^2\left(\frac{2Tt}{\delta}\right) $$

For \textbf{E2}, we similarly require the norm bounds on $y_{t-1}$ to hold for all $T$ time steps. By setting the failure probability to $\frac{\delta}{2T}$ at each step $t$ within the proof of Lemma 2, the union bound ensures that all such bounds hold simultaneously with a total failure probability of at most $T \times \frac{\delta}{2T} = \delta/2$.

By the union bound, the total probability of failure for any of these events across all time is at most $P(\neg E1) + P(\neg E2) \le \delta/2 + \delta/2 = \delta$. Therefore, all required bounds hold simultaneously with a probability of at least $1-\delta$. The theorem Statement follows by combining the results from Sections \ref{lemma:instantaneous_regret_decomposition} and \ref{lemma:cumulative_regret} under this high-probability event.
\hfill\ensuremath{}

\end{proof}

\section{Regret Analysis for Additive Compositional Kernels}
\label{sec:regret_compositional}

This section extends the preceding analysis to a specific, structured kernel learning scenario. We assume the true kernel $k^*$ is an additive composition of two stationary kernels, and the algorithm learns a kernel $k_{t-1}$ that mirrors this additive structure. Our goal is to derive a regret bound that explicitly depends on the kernel mismatch of the individual ``child'' components.

\subsection{Problem Setup and Notation}

We adopt the problem setup from Section \ref{sec:regret_learned} with the following structural specializations.
\begin{description}
    \item[Kernel Structure:] The true kernel $k^*$ is assumed to be the sum of two stationary kernels:
    \[
    k^*(x, x') = k^*_1(x, x') + k^*_2(x, x')
    \]
    The algorithm is aware of this structure and employs a learned kernel $k_{t-1}$ at each step $t$ that is also a sum of learned components:
    \[
    k_{t-1}(x, x') = k_{1, t-1}(x, x') + k_{2, t-1}(x, x')
    \]
    \item[Component-wise Kernel Mismatch:] We define the uniform deviation for each ``child'' kernel component $i \in \{1, 2\}$ at step $t$ as:
    \[
    \Delta_{i, t} := \sup_{x, x' \in \mathcal{X}} |k_{i, t-1}(x, x') - k^*_i(x, x')|
    \]
    This is in contrast to the total kernel mismatch $\Delta_t$ defined in the original analysis.
    
\end{description}

\subsection{Assumptions}

We rely on Assumptions 1 and 2 from the original analysis (Section \ref{sec:regret_learned}, with a necessary clarification for the additive structure of Assumption 2.

\textbf{Assumption 1 (Bounded RKHS Norm):} This assumption remains unchanged. The true function $f$ has a bounded norm $||f||_{\mathcal{H}_{k^*}} \le B$. Note that for an additive kernel, the RKHS $\mathcal{H}_{k^*}$ is the direct sum of the component RKHSs ($\mathcal{H}_{k^*_1} \oplus \mathcal{H}_{k^*_2}$), and the norm is
\[
||f||^2_{\mathcal{H}_{k^*}} = \inf_{f=f_1+f_2} (||f_1||^2_{\mathcal{H}_{k^*_1}} + ||f_2||^2_{\mathcal{H}_{k^*_2}}).
\]

\textbf{Assumption 2 (Additive Bounded Kernel Variance):} This is a structural extension of Assumption from section \ref{sec:regret_learned}. We assume the diagonal of both the true and learned component kernels are uniformly bounded. There exist constants $\kappa_1, \kappa_2 > 0$ such that for all $t \in \{1, \dots, T\}$ and $i \in \{1, 2\}$:
\[
\sup_{x \in \mathcal{X}} k^*_i(x, x) \le \kappa_i^2
\]
\[
\sup_{x \in \mathcal{X}} k_{i, t-1}(x, x) \le \kappa_i^2
\]
This implies the bound for the total kernels holds with $\kappa^2 = \kappa_1^2 + \kappa_2^2$, as:
\[
\sup_{x \in \mathcal{X}} k^*(x, x) = \sup_{x \in \mathcal{X}} (k^*_1(x, x) + k^*_2(x, x)) \le \kappa_1^2 + \kappa_2^2 = \kappa^2
\]
\[
\sup_{x \in \mathcal{X}} k_{t-1}(x, x) = \sup_{x \in \mathcal{X}} (k_{1, t-1}(x, x) + k_{2, t-1}(x, x)) \le \kappa_1^2 + \kappa_2^2 = \kappa^2
\]
Therefore, all proofs from the preceding sections that rely on a uniform bound $\kappa^2$ (and its square root $\kappa$) remain valid.

\subsection{Bound on Total Kernel Mismatch}

We first establish a simple lemma that bounds the total kernel mismatch $\Delta_t$ as a function of the component mismatches $\Delta_{1, t}$ and $\Delta_{2, t}$.

\begin{lemma}(Compositional Mismatch Bound).
\label{lemma:compositional_mismatch}
    
Under the additive kernel setup, the total kernel mismatch $\Delta_t$ is bounded by the sum of the component kernel mismatches:
\[
\Delta_t \le \Delta_{1, t} + \Delta_{2, t}
\]
\end{lemma}
\begin{proof}
We begin with the definition of the total kernel mismatch $\Delta_t$:
\[
\Delta_t = \sup_{x, x' \in \mathcal{X}} |k_{t-1}(x, x') - k^*(x, x')|
\]
Substitute the additive definitions of $k_{t-1}$ and $k^*$:
\[
\Delta_t = \sup_{x, x' \in \mathcal{X}} |(k_{1, t-1}(x, x') + k_{2, t-1}(x, x')) - (k^*_1(x, x') + k^*_2(x, x'))|
\]
Rearrange the terms inside the absolute value:
\[
\Delta_t = \sup_{x, x' \in \mathcal{X}} |(k_{1, t-1}(x, x') - k^*_1(x, x')) + (k_{2, t-1}(x, x') - k^*_2(x, x'))|
\]
Apply the triangle inequality, $|a+b| \le |a| + |b|$, for any $x, x' \in \mathcal{X}$:
\[
|(k_{1, t-1} - k^*_1) + (k_{2, t-1} - k^*_2)| \le |k_{1, t-1}(x, x') - k^*_1(x, x')| + |k_{2, t-1}(x, x') - k^*_2(x, x')|
\]
The supremum of a sum is less than or equal to the sum of the suprema:
\[
\Delta_t \le \sup_{x, x' \in \mathcal{X}} |k_{1, t-1}(x, x') - k^*_1(x, x')| + \sup_{x, x' \in \mathcal{X}} |k_{2, t-1}(x, x') - k^*_2(x, x')|
\]
By the definition of the component-wise mismatch (Assumption 2), this gives:
\[
\Delta_t \le \Delta_{1, t} + \Delta_{2, t}
\]
\end{proof}

\subsection{Corollary for Additive Kernels}

We can now State the main regret bound as a corollary of Theorem \ref{thm:main_theorem}. We directly substitute the bound from Lemma \ref{lemma:compositional_mismatch} into the perturbation bounds $\delta_{\mu,t}$ and $\delta_{\sigma,t}$ derived in Lemmas \ref{lemma:posterior_mean_perturbation} and \ref{lemma:posterior_variance_perturbation}.

To make the dependency explicit, let us define $\hat{\delta}_{\mu,t}$ and $\hat{\delta}_{\sigma,t}$ as upper bounds on the original perturbation terms, based on the component mismatches.

From Lemma \ref{lemma:posterior_mean_perturbation}, the posterior mean perturbation is bounded by $\delta_{\mu,t}$:
\[
\delta_{\mu,t} = \frac{\sqrt{t-1}\Delta_{t}}{\sigma^{2}}(B\kappa+\sigma\sqrt{t-1}+\sigma\sqrt{2\log(\tfrac{2T}{\delta})})(1+\tfrac{\kappa(t-1)}{\sigma^{2}})
\]
Applying Lemma \ref{lemma:compositional_mismatch}, we have $\Delta_t \le \Delta_{1, t} + \Delta_{2, t}$. We can thus define a new, explicit bound $\hat{\delta}_{\mu,t}$:

\begin{center}
    \begin{equation}
    \delta_{\mu,t} \le \frac{\sqrt{t-1}(\Delta_{1, t} + \Delta_{2, t})}{\sigma^{2}}(B\kappa+\sigma\sqrt{t-1}+\sigma\sqrt{2\log(\tfrac{2T}{\delta})})(1+\tfrac{\kappa(t-1)}{\sigma^{2}}) := \hat{\delta}_{\mu,t}
    \end{equation}
\end{center}

Similarly, from Lemma \ref{lemma:posterior_variance_perturbation}, the posterior variance perturbation is bounded by $\delta_{\sigma,t}$:
\[
\delta_{\sigma,t} = \frac{\kappa(t-1)\Delta_{t}}{\sigma^{2}}\left(\frac{\kappa(t-1)}{\sigma^{2}}+2\right)
\]
Applying Lemma \ref{lemma:compositional_mismatch}, we define the explicit bound $\hat{\delta}_{\sigma,t}$:
\begin{center}
    \begin{equation}
        \delta_{\sigma,t} \le \frac{\kappa(t-1)(\Delta_{1, t} + \Delta_{2, t})}{\sigma^{2}}\left(\frac{\kappa(t-1)}{\sigma^{2}}+2\right) := \hat{\delta}_{\sigma,t}
    \end{equation}
\end{center}
We now substitute these new bounds, $\hat{\delta}_{\mu,t}$ and $\hat{\delta}_{\sigma,t}$, into the final regret bound from Theorem \ref{thm:main_theorem}.

\begin{corollary} Regret Bound for GP-UCB with a Learned Hierarchical Kernel
\label{thm:regret_structured}

Under Assumptions 1 and the Additive version of Assumption 2, let $\delta \in (0, 1)$ and set $\beta_t$ as defined in the proof of Theorem \ref{thm:main_theorem}. With probability at least $1-\delta$, the cumulative regret $R_T$ of the GP-UCB algorithm using the structured kernel $k_{t-1} = k_{1, t-1} + k_{2, t-1}$ is bounded by:
\[
R_{T} \le 2\sqrt{T\beta_{T}(C\gamma_{T}(k^{*}) + \sum_{t=1}^{T}\hat{\delta}_{\sigma,t})} + \sum_{t=1}^{T}(2\hat{\delta}_{\mu,t} + 2\sqrt{\beta_{t}\hat{\delta}_{\sigma,t}})
\]
where $C=\frac{2}{\log(1+\sigma^{-2})}$, $\beta_{T}=\max_{t \le T}\beta_{t}$, $\gamma_T(k^*)$ is the maximum information gain with the true kernel, and the perturbation bounds $\hat{\delta}_{\mu,t}$ and $\hat{\delta}_{\sigma,t}$ depend explicitly on the component-wise kernel mismatches $\Delta_{1, t}$ and $\Delta_{2, t}$ as:
\[
\hat{\delta}_{\mu,t} := \frac{\sqrt{t-1}(\Delta_{1, t} + \Delta_{2, t})}{\sigma^{2}}(B\kappa+\sigma\sqrt{t-1}+\sigma\sqrt{2\log(\tfrac{2T}{\delta})})(1+\tfrac{\kappa(t-1)}{\sigma^{2}})
\]
\[
\hat{\delta}_{\sigma,t} := \frac{\kappa(t-1)(\Delta_{1, t} + \Delta_{2, t})}{\sigma^{2}}\left(\frac{\kappa(t-1)}{\sigma^{2}}+2\right)
\]
\end{corollary}

\begin{proof} The proof follows directly from Theorem \ref{thm:main_theorem}. The probabilistic events and union bound arguments remain identical. The instantaneous regret decomposition $r_t \le 2\sqrt{\beta_{t}}\sigma_{t-1}(x_{t})+2\delta_{\mu,t}+2\sqrt{\beta_{t}\delta_{\sigma,t}}$ holds, leading to the cumulative regret bound $R_{T} \le 2\sqrt{T\beta_{T}(C\gamma_{T}(k^{*}) + \sum_{t=1}^{T}\hat{\delta}_{\sigma,t})} + \sum_{t=1}^{T}(2\hat{\delta}_{\mu,t} + 2\sqrt{\beta_{t}\hat{\delta}_{\sigma,t}})$. 

By Lemma \ref{lemma:compositional_mismatch}, we have $\Delta_t \le \Delta_{1, t} + \Delta_{2, t}$. The perturbation bounds $\delta_{\mu,t}$ and $\delta_{\sigma,t}$ are defined in Lemmas \ref{lemma:posterior_mean_perturbation} and \ref{lemma:posterior_variance_perturbation} as products of $\Delta_t$ and other non-negative terms. Therefore, both bounds are monotonically non-decreasing in $\Delta_t$. We can substitute $\Delta_t$ with its upper bound $\Delta_{1, t} + \Delta_{2, t}$ to yield the new, explicit bounds $\hat{\delta}_{\mu,t} \ge \delta_{\mu,t}$ and $\hat{\delta}_{\sigma,t} \ge \delta_{\sigma,t}$. Substituting these upper bounds into the final regret inequality of Theorem \ref{thm:main_theorem} preserves the inequality and generalizes it to the new kernel structure $k_{t-1} = k_{1,t-1} + k_{2,t-1}$.

\end{proof}
This theorem explicitly demonstrates that the regret bound is driven by the sum of the estimation errors of the component kernels. If the component mismatches $\Delta_{1, t}$ and $\Delta_{2, t}$ diminish over time, so too will the regret attributable to kernel misspecification.

\section{BIF Mathematical Characterization}
\label{app:prop_proof}

Theorem \ref{main_thm:regret_bound} and Corollary \ref{main_cor:hierarchical_regret} establish that controlling the kernel mismatch $\Delta_t$ is sufficient for bounding cumulative regret. We now characterize BIF's two information channels: the upward prior map and the downward response contribution. Proposition~\ref{prop:prior_map} shows the prior map converges uniformly to the true objective as children improve. Proposition \ref{prop:contribution_bias} characterizes the bias introduced by the softmax contribution decomposition. 

\subsection{Prior Map Convergence}

\begin{definition}[Child Prediction Quality]\label{def:child_quality}
At step $t$, define the prediction error and residual exploration
of child $s$ as:
\begin{align}
    \varepsilon_{s,t} &:= \sup_{\mathbf{x}_s \in \mathcal{X}_s}
        |\mu_{s,t}(\mathbf{x}_s) - g_s^*(\mathbf{x}_s)|,
        \label{eq:pred_error}\\
    \delta^{\mathrm{exp}}_{s,t} &:= \sup_{\mathbf{x}_s \in \mathcal{X}_s}
        \frac{\sigma_{s,t}(\mathbf{x}_s)}
             {\sqrt{n_{s,t}(\mathbf{x}_s) \vee 1}},
        \label{eq:resid_exp}
\end{align}
where $g_1^* = g^*$, $g_2^* = h^*$,\;
$\mu_{s,t}$ and $\sigma_{s,t}$ are the posterior mean and standard
deviation of child~$s$ at step~$t$,\;
$\gamma > 0$ is the child exploration weight
(Algorithm~\ref{alg:BIF}, line~3;\@ Eq.~\ref{def:response_contribution}),\;
$a \vee b := \max(a,b)$ ensures the quantity is well-defined at
unqueried grid points (the floor is vacuous once every point has been
queried at least once),\;
$\mathcal{X}_s$ is the finite search grid used by child~$s$,\;
and $\mathcal{X} = \mathcal{X}_1 \times \cdots \times \mathcal{X}_{|S|}$
is the parent search grid.
\end{definition}

\begin{proposition}[Prior Map Convergence]\label{prop:prior_map}
Under the quasi-linear additive assumption
$f^*(\mathbf{x}) = \sum_{s=1}^{|S|} g_s^*(\mathbf{x}_s)$,
the BIF prior map
$A_t(\mathbf{x}) = \frac{1}{|S|}\sum_{s=1}^{|S|} \alpha_{s,t}(\mathbf{x}_s)$
satisfies, for all $\mathbf{x} \in \mathcal{X}$ and $t \geq 1$:
\begin{equation}\label{eq:prior_map_bound}
    \sup_{\mathbf{x} \in \mathcal{X}}
    \left| |S| \cdot A_t(\mathbf{x}) - f^*(\mathbf{x}) \right|
    \leq \sum_{s=1}^{|S|} \varepsilon_{s,t}
       + \gamma \sum_{s=1}^{|S|} \delta^{\mathrm{exp}}_{s,t}.
\end{equation}
In particular, for $|S| = 2$:
\[
    \sup_{\mathbf{x}} |2A_t(\mathbf{x}) - f^*(\mathbf{x})|
    \leq (\varepsilon_{1,t} + \varepsilon_{2,t})
       + \gamma(\delta^{\mathrm{exp}}_{1,t} + \delta^{\mathrm{exp}}_{2,t}).
\]
As child models improve
($\varepsilon_{s,t}, \delta^{\mathrm{exp}}_{s,t} \to 0$),
the prior map converges uniformly to $f^*/|S|$.
\end{proposition}

\begin{proof}
\label{proof:prior_map_convergence}
By definition of the BIF prior map (Algorithm~\ref{alg:BIF}, lines~3--4):
\[
    |S| \cdot A_t(\mathbf{x})
    = \sum_{s=1}^{|S|} \alpha_{s,t}(\mathbf{x}_s)
    = \sum_{s=1}^{|S|} \left[ \mu_{s,t}(\mathbf{x}_s)
    + \gamma \cdot
      \frac{\sigma_{s,t}(\mathbf{x}_s)}
           {\sqrt{n_{s,t}(\mathbf{x}_s) \vee 1}} \right].
\]
Under the additive assumption,
$f^*(\mathbf{x}) = \sum_{s=1}^{|S|} g_s^*(\mathbf{x}_s)$.
Therefore:
\begin{align*}
    \left| |S| \cdot A_t(\mathbf{x}) - f^*(\mathbf{x}) \right|
    &= \left| \sum_{s=1}^{|S|} \left[
       \mu_{s,t}(\mathbf{x}_s) - g_s^*(\mathbf{x}_s)
       + \gamma \cdot
         \frac{\sigma_{s,t}(\mathbf{x}_s)}
              {\sqrt{n_{s,t}(\mathbf{x}_s) \vee 1}}
       \right] \right| \\
    &\leq \sum_{s=1}^{|S|}
       |\mu_{s,t}(\mathbf{x}_s) - g_s^*(\mathbf{x}_s)|
       + \gamma \sum_{s=1}^{|S|}
         \frac{\sigma_{s,t}(\mathbf{x}_s)}
              {\sqrt{n_{s,t}(\mathbf{x}_s) \vee 1}} \\
    &\leq \sum_{s=1}^{|S|} \varepsilon_{s,t}
       + \gamma \sum_{s=1}^{|S|} \delta^{\mathrm{exp}}_{s,t}.
\end{align*}
The first inequality is the triangle inequality.
The second follows from
Definitions~\eqref{eq:pred_error} and~\eqref{eq:resid_exp}.
\end{proof}

\subsection{Contribution Bias}

\begin{proposition}[Contribution Bias Characterization]\label{prop:contribution_bias}
Under the additive assumption, let
$y = g^*(\mathbf{x}_1) + h^*(\mathbf{x}_2)$ be the parent response
and $y_1 = y \cdot w_1$ the BIF contribution to child~1
(Definition~\ref{def:response_contribution}), where
$w_1 = \frac{e^{c_1}}{e^{c_1} + e^{c_2}}$ with $c_s$ as in Eq.~\ref{eq:softmax}.
Define the ideal weight
$w_1^* = \frac{g^*(\mathbf{x}_1)}{g^*(\mathbf{x}_1) + h^*(\mathbf{x}_2)}$
for $g^*(\mathbf{x}_1) > 0$ and $h^*(\mathbf{x}_2) > 0$.
Then the contribution bias
$b_1(\mathbf{x}) := y_1 - g^*(\mathbf{x}_1)$ satisfies:
\begin{enumerate}
    \item \textbf{Decomposition.}\;
    $b_1(\mathbf{x}) = f^*(\mathbf{x}) \cdot (w_1 - w_1^*)$.

    \item \textbf{Lipschitz bound.}\;
    $\displaystyle
    |b_1(\mathbf{x})| \leq \frac{|f^*(\mathbf{x})|}{4}
    \left| (c_1 - c_2)
    - \log\frac{g^*(\mathbf{x}_1)}{h^*(\mathbf{x}_2)} \right|.$

    \item \textbf{Balanced case.}\;
    If\; $g^*(\mathbf{x}_1) = h^*(\mathbf{x}_2)$
    and $\sup_{\mathbf{x}_1} g^*(\mathbf{x}_1) = \sup_{\mathbf{x}_2} h^*(\mathbf{x}_2) > 0$,
    then $b_1(\mathbf{x}) \to 0$ for each such $\mathbf{x}$, as
    $\varepsilon_{s,t}, \delta^{\mathrm{exp}}_{s,t} \to 0$.
\end{enumerate}
\end{proposition}
    
\label{proof:contribution_bias}

\paragraph{Part (1): Decomposition.}
\begin{proof}
    Since $g^* > 0$ and $h^* > 0$, we have $f^* = g^* + h^* > 0$ and:
\begin{align*}
    b_1(\mathbf{x})
    &= y \cdot w_1 - g^*(\mathbf{x}_1)
     = (g^* + h^*) \cdot w_1 - g^* \\
    &= (g^* + h^*) \cdot
       \left(w_1 - \frac{g^*}{g^* + h^*}\right)
     = f^*(\mathbf{x}) \cdot (w_1 - w_1^*).
\end{align*}
\end{proof}

\paragraph{Part (2): Lipschitz bound.}
\begin{proof}
    
Since $g^* > 0$ and $h^* > 0$, the ideal weight admits the
sigmoid representation
$w_1^* = \varsigma\!\left(\log\frac{g^*}{h^*}\right)$,
where $\varsigma(z) = \frac{1}{1+e^{-z}}$ is the sigmoid function
and the BIF weight is $w_1 = \varsigma(c_1 - c_2)$.

The sigmoid satisfies
$|\varsigma'(z)| = \varsigma(z)(1 - \varsigma(z)) \leq \frac{1}{4}$
for all $z \in \mathbb{R}$,
with equality at $z = 0$.
By the mean value theorem:
\[
    |w_1 - w_1^*|
    = |\varsigma(c_1 - c_2) - \varsigma(\log(g^*/h^*))|
    \leq \frac{1}{4}
    \left| (c_1 - c_2) - \log\frac{g^*}{h^*} \right|.
\]
Multiplying by $|f^*(\mathbf{x})|$ and applying part~(1)
yields the result.
\end{proof}

\paragraph{Part (3): Balanced case.}

\begin{proof}
When $g^*(\mathbf{x}_1) = h^*(\mathbf{x}_2)$,
we have $w_1^* = \frac{1}{2}$,\; i.e., $\log(g^*/h^*) = 0$.

As children converge
($\varepsilon_{s,t} \to 0$ and
$\delta^{\mathrm{exp}}_{s,t} \to 0$,
with $\sup_{\mathbf{x}_s} g_s^*(\mathbf{x}_s) > 0$ for each $s$),
the denominator
$\sup_{\tilde{\mathbf{x}}_s} \alpha_{s,t}(\tilde{\mathbf{x}}_s)$
is eventually positive, and the contribution scores satisfy
$c_s \to g_s^*(\mathbf{x}_s) /
  \sup_{\tilde{\mathbf{x}}_s} g_s^*(\tilde{\mathbf{x}}_s)$.
Under the condition
$\sup_{\mathbf{x}_1} g^*(\mathbf{x}_1) = \sup_{\mathbf{x}_2} h^*(\mathbf{x}_2)$ and $g^*(\mathbf{x}_1) = h^*(\mathbf{x}_2)$,
this gives $c_1 - c_2 \to 0$, hence
$w_1 \to \frac{1}{2} = w_1^*$ and $b_1(\mathbf{x}) \to 0$
by part~(1).
\end{proof}

\subsection{Stability Properties of Response Contribution System}
\label{app:stability_properties}

\paragraph{Stability Properties.}
Three properties follow directly from the response contribution system outlined in Section \ref{def:response_contribution}:
\begin{enumerate}
    \item \textbf{Exploration-Guided Regularization.} When a child $s$ is data-starved, its predictive variance $\sigma_s(x)$ inflates relative to $\sqrt{n_s(x)}$, inflating $u_s(x)$ and therefore $w_s(x)$. This self-correcting behaviour directs a larger share of the global signal toward under-informed children until their uncertainty collapses, preventing any child from stagnating post-initialization.

    \item \textbf{Bounded Logits.} Stage~1 normalization guarantees $c_s(x) \in (0,\,1]$, so the softmax inputs are uniformly bounded independent of the magnitude of $\mu_s$ or $\sigma_s$, ensuring numerical stability throughout training.

    \item \textbf{Conservative Signal Partitioning.} The simplex constraint $\sum_{s \in S} w_s(x) = 1$ ensures that no additional energy is injected into the hierarchy; each child receives a strict fraction of the observed signal. This is essential for stable marginal likelihood optimization of the child kernels $\theta_s$, as duplicating the signal would introduce a systematic positive bias into the noise estimates.
\end{enumerate}

\section*{NeurIPS Paper Checklist}

\begin{enumerate}

\item {\bf Claims}
    \item[] Question: Do the main claims made in the abstract and introduction accurately reflect the paper's contributions and scope?
    \item[] Answer: \answerYes{} 
    \item[] Justification: All claims in the abstract are supported by either empirical or mathematically proven theoretical evidence.

\item {\bf Limitations}
    \item[] Question: Does the paper discuss the limitations of the work performed by the authors?
    \item[] Answer: \answerYes{} 
    \item[] Justification: Limitations discussed in the conclusion about runtime and structural assumptions

\item {\bf Theory assumptions and proofs}
    \item[] Question: For each theoretical result, does the paper provide the full set of assumptions and a complete (and correct) proof?
    \item[] Answer: \answerYes{} 
    \item[] Justification: All proofs are provided with assumptions made in the appendix

    \item {\bf Experimental result reproducibility}
    \item[] Question: Does the paper fully disclose all the information needed to reproduce the main experimental results of the paper to the extent that it affects the main claims and/or conclusions of the paper (regardless of whether the code and data are provided or not)?
    \item[] Answer: \answerYes{} 
    \item[] Justification: code, datasets, and resources required are provided in the maintext and appendix.

\item {\bf Open access to data and code}
    \item[] Question: Does the paper provide open access to the data and code, with sufficient instructions to faithfully reproduce the main experimental results, as described in supplemental material?
    \item[] Answer: \answerYes{} 
    \item[] Justification: All code used to run the experiments have been provided as supplementary material

\item {\bf Experimental setting/details}
    \item[] Question: Does the paper specify all the training and test details (e.g., data splits, hyperparameters, how they were chosen, type of optimizer) necessary to understand the results?
    \item[] Answer: \answerYes{} 
    \item[] Justification: all optimization landscapes are clearly stated with hyperparameters for the models given.

\item {\bf Experiment statistical significance}
    \item[] Question: Does the paper report error bars suitably and correctly defined or other appropriate information about the statistical significance of the experiments?
    \item[] Answer: \answerYes{} 
    \item[] Justification: All experiments are run on 10 repitions with standard error reported in the tables and figures

\item {\bf Experiments compute resources}
    \item[] Question: For each experiment, does the paper provide sufficient information on the computer resources (type of compute workers, memory, time of execution) needed to reproduce the experiments?
    \item[] Answer: \answerYes{} 
    \item[] Justification: Compute resources used as well as runtimes are given in the appendix.
    
\item {\bf Code of ethics}
    \item[] Question: Does the research conducted in the paper conform, in every respect, with the NeurIPS Code of Ethics \url{https://neurips.cc/public/EthicsGuidelines}?
    \item[] Answer: \answerYes{} 
    \item[] Justification: All ethical guidelines are respected

\item {\bf Broader impacts}
    \item[] Question: Does the paper discuss both potential positive societal impacts and negative societal impacts of the work performed?
    \item[] Answer: \answerNA{} 
    \item[] Justification: The proposed solutions and datasets provide no further impact other than scientific novelty.
    
\item {\bf Safeguards}
    \item[] Question: Does the paper describe safeguards that have been put in place for responsible release of data or models that have a high risk for misuse (e.g., pre-trained language models, image generators, or scraped datasets)?
    \item[] Answer: \answerNA{} 
    \item[] Justification: No safeguards must be put in place as is not required for other optimization algorithms

\item {\bf Licenses for existing assets}
    \item[] Question: Are the creators or original owners of assets (e.g., code, data, models), used in the paper, properly credited and are the license and terms of use explicitly mentioned and properly respected?
    \item[] Answer: \answerYes{} 
    \item[] Justification: All authors are credited appropriately with licenses given.

\item {\bf New assets}
    \item[] Question: Are new assets introduced in the paper well documented and is the documentation provided alongside the assets?
    \item[] Answer: \answerYes{} 
    \item[] Justification: All novel code and datasets are properly credited to the respective authors.

\item {\bf Crowdsourcing and research with human subjects}
    \item[] Question: For crowdsourcing experiments and research with human subjects, does the paper include the full text of instructions given to participants and screenshots, if applicable, as well as details about compensation (if any)? 
    \item[] Answer: \answerNA{} 
    \item[] Justification: No crowdsourcing or human subjects were used in this work.

\item {\bf Institutional review board (IRB) approvals or equivalent for research with human subjects}
    \item[] Question: Does the paper describe potential risks incurred by study participants, whether such risks were disclosed to the subjects, and whether Institutional Review Board (IRB) approvals (or an equivalent approval/review based on the requirements of your country or institution) were obtained?
    \item[] Answer: \answerNA{} 
    \item[] Justification: No crowdsourcing or human subjects were used in this work.

\item {\bf Declaration of LLM usage}
    \item[] Question: Does the paper describe the usage of LLMs if it is an important, original, or non-standard component of the core methods in this research? Note that if the LLM is used only for writing, editing, or formatting purposes and does \emph{not} impact the core methodology, scientific rigor, or originality of the research, declaration is not required.
    \item[] Answer: \answerNA{} 
    \item[] Justification: LLMs were only used for editing and formatting purposes.

\end{enumerate}

\end{document}